\documentclass[journal]{IEEEtran}
\usepackage{cite}
\usepackage{amsmath,amssymb,amsfonts}
\usepackage{algorithm}
\usepackage{algorithmic}
\usepackage{graphicx}
\usepackage{textcomp}
\usepackage{epstopdf}
\usepackage{amsmath}
\usepackage{amssymb}
\usepackage{algorithm}
\usepackage{algorithmic}
\usepackage{multirow}
\usepackage{amsmath}
\usepackage{xcolor}
\usepackage{graphicx}
\usepackage{subfigure}
\usepackage{epsfig}
\usepackage{epstopdf}
\usepackage{multirow}
\usepackage{array}
\usepackage{url}
\usepackage{hyperref}
\usepackage{multirow}
\usepackage{booktabs}
\usepackage{makecell}
\hypersetup{
    bookmarks=true,         % show bookmarks bar?
    unicode=false,          % non-Latin characters in Acrobat's bookmarks
    pdftoolbar=true,        % show Acrobat's toolbar?
    pdfmenubar=true,        % show Acrobat's menu?
    pdffitwindow=false,     % window fit to page when opened
    pdfstartview={FitH},    % fits the width of the page to the window
    pdftitle={My title},    % title
    pdfauthor={Author},     % author
    pdfsubject={Subject},   % subject of the document
    pdfcreator={Creator},   % creator of the document
    pdfproducer={Producer}, % producer of the document
    pdfkeywords={keyword1, key2, key3}, % list of keywords
    pdfnewwindow=true,      % links in new PDF window
    colorlinks=true,       % false: boxed links; true: colored links
    linkcolor=black,          % color of internal links (change box color with linkbordercolor)
    citecolor=black,        % color of links to bibliography
    filecolor=black,      % color of file links
    urlcolor=black           % color of external links
}
\newcommand{\PreserveBackslash}[1]{\let\temp=\\#1\let\\=\temp}
\newcolumntype{C}[1]{>{\PreserveBackslash\centering}p{#1}}
\newcolumntype{R}[1]{>{\PreserveBackslash\raggedleft}p{#1}}
\newcolumntype{L}[1]{>{\PreserveBackslash\raggedright}p{#1}}

\newcommand{\xiaosihao}{\fontsize{10pt}{\baselineskip}\selectfont}

\begin{document}
\title{Adaptively Meshed Video Stabilization}

\author{Minda Zhao, Qiang Ling,~\IEEEmembership{Senior Member,~IEEE}
\vspace{-0.1in}
       % <-this % stops a space
\thanks{ M. Zhao, Q. Ling are with Dept. of Automation, University of Science and Technology of China, Hefei, Anhui 230027, P. R. China; Email:
qling@ustc.edu.cn}% <-this % stops a space
}
\maketitle

\begin{abstract}
Video stabilization is essential for improving visual quality of shaky videos. The current video stabilization methods usually take feature trajectories in the background to estimate one global transformation matrix or several transformation matrices based on a fixed mesh, and warp shaky frames into their stabilized views. However, these methods may not model the shaky camera motion well in complicated scenes, such as scenes containing large foreground objects or strong parallax, and may result in notable visual artifacts in the stabilized videos. To resolve the above issues, this paper proposes an adaptively meshed method to stabilize a shaky video based on all of its feature trajectories and an adaptive blocking strategy. More specifically, we first extract feature trajectories of the shaky video and then generate a triangle mesh according to the distribution of the feature trajectories in each frame. Then transformations between shaky frames and their stabilized views over all triangular grids of the mesh are calculated to stabilize the shaky video. Since more feature trajectories can usually be extracted from all regions, including both background and foreground regions, a finer mesh will be obtained and provided for camera motion estimation and frame warping. We estimate the mesh-based transformations of each frame by solving a two-stage optimization problem. Moreover, foreground and background feature trajectories are no longer distinguished and both contribute  to the estimation of the camera motion in the proposed optimization problem, which yields better estimation performance than previous works, particularly in challenging videos with large foreground objects or strong parallax. To further enhance the robustness of our method, we propose two adaptive weighting mechanisms to improve its spatial and temporal adaptability. Experimental results demonstrate the effectiveness of our method in producing visually pleasing stabilization effects in various challenging videos.

\end{abstract}


\IEEEpeerreviewmaketitle

%\vspace{-0.1in}
\section{Introduction}

\IEEEPARstart {W}ith the popularity of portable camera devices like digital phones, people get used to recording daily lives and important events, such as birthdays and weddings, into videos\cite{joshi2015real}, \cite{liu2009content}. Due to the lack of professional stabilizing instruments, the recorded amateur videos may be shaky with undesirable jitters, which yield unpleasant visual discomfort. Serious video shakiness may also degrade the performance of subsequent computer vision tasks, such as tracking\cite{gard2019projection}, video action recognition\cite{wang2016temporal} and video classification\cite{liu2016towards}. So video stabilization becomes an essential task for shaky videos.

\begin{figure}[!hbt]
\centering\includegraphics[width=\hsize]{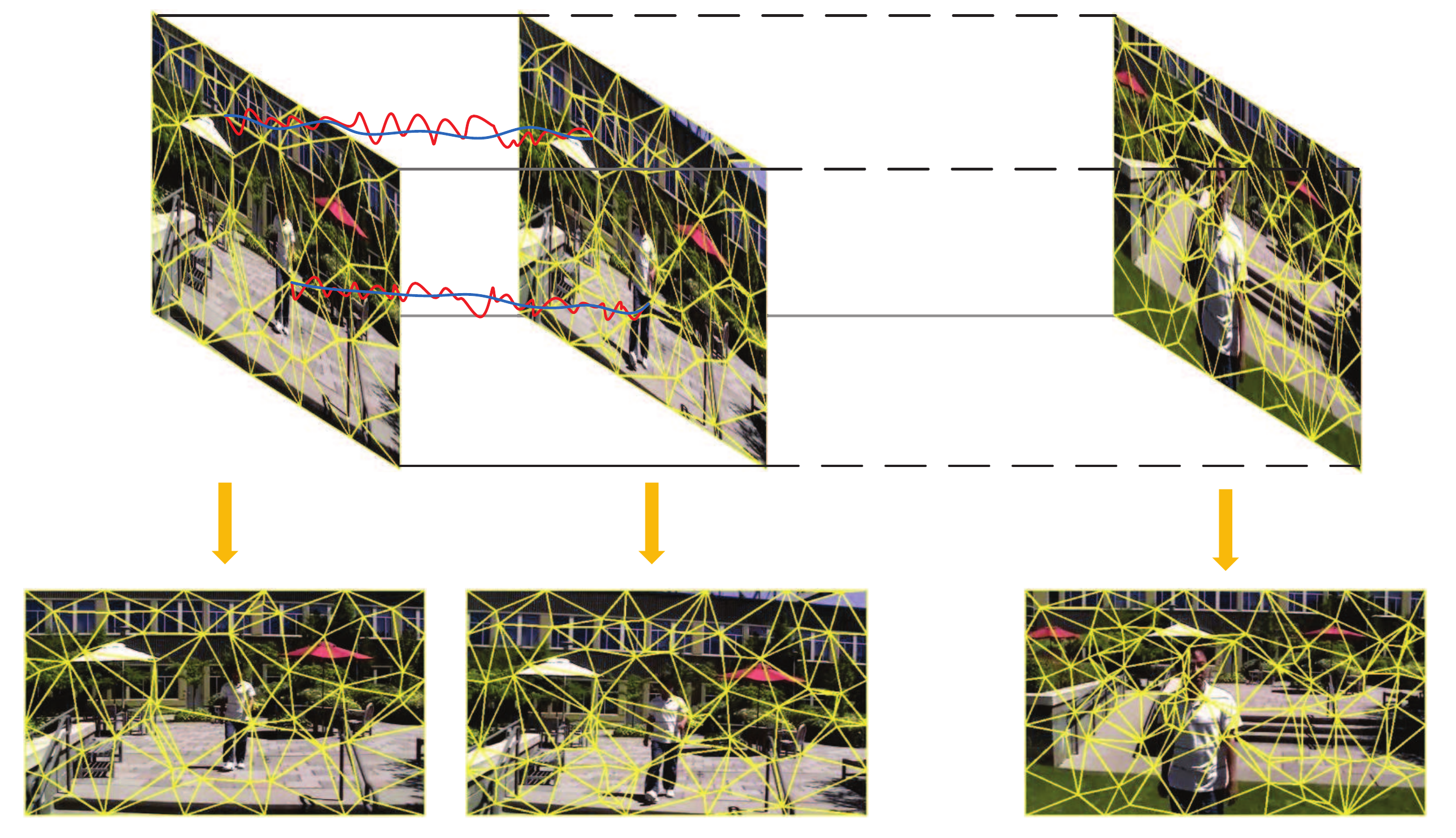}
\caption{Triangle meshes of several frames in an example video. The first row shows the procedure of generating triangle meshes according to feature trajectories.  The red curves stand for the original feature trajectories and the smoothed ones are represented with blue curves. The second row shows the obtained triangle meshes. Since the first two frames are close temporally, their triangle meshes are more similar while the triangle meshes of the last two frames are quite different because of their weak temporal continuity.}
\label{fig:description}
\end{figure}

Traditional video stabilization methods can be roughly categorized into 2D methods, 3D methods and 2.5D methods. 2D methods usually model the camera motion with inter-frame transformation matrices and then smooth these matrices elaborately \cite{chen2008capturing}, \cite{gleicher2008re}, \cite{morimoto1998evaluation}. To enhance the robustness of the 2D stabilization methods, some techniques introduce the {\it ``as-similar-as-possible''} constraint \cite{zhang2018intrinsic} and divide the whole frame into a fixed grid mesh \cite{wang2018deep}. These techniques are computationally efficient, but fragile in the existence of parallax. 3D methods are robust against parallax, which track a set of feature points and reconstruct their 3D locations by the structure-from-motion(SFM) technique\cite{hartley2003multiple}. However, 3D methods are computationally expensive and are not as robust as 2D methods. Recently, combining the advantages of 2D methods and 3D methods, 2.5D methods emerge. 2.5D methods first extract feature points and match them frame by frame to generate feature trajectories, then smooth these feature trajectories, and finally warp frames based on the feature point pairs of the original and stabilized views. When warping frames to their stabilized views, one global transformation matrix or several mesh-based transformation matrices \cite{koh2015video} are calculated. The proposed method of the present paper also belongs to 2.5D methods.

Although the available methods can stabilize most videos well, they may suffer performance degradation in some complicated scenes.
\begin{itemize}
\item On the one hand,  foreground objects with different movements often appear in the same frame and the inter-frame motion of different objects can not be simply represented by the same transformation, either the same affine matrix or the same homography. Some methods were proposed to solve this problem by dividing a frame into a fixed grid mesh and estimating a transformation matrix between the shaky frame and its stabilized view for each grid of the mesh. However, they ignore the contents of the video frames and consider all grids equally. Typically, a region lacking in information, such as roads and the sky, and a region containing a crowd of people are divided into grids of the same size, and are expected to perform the same transformation inside each grid. The former grid may be too small since the same transformation can work well in a larger region. However, the latter grid may be too large because it contains discontinuous depth variation among different objects and should be divided into finer grids. Thus such fixed mesh-based video stabilization methods may not work well in some complicated scenes.
\item On the other hand, many stabilization methods discard foreground feature trajectories and estimate the camera motion with only background feature trajectories. So they warp foreground regions only under the guidance of nearby background regions. This may result in distortion in the foreground regions. Furthermore, it is not an easy task to accurately determine background feature trajectories,  especially in complicated scenes, such as a scene with large foreground objects.
\end{itemize}

To resolve the above issues, we propose a novel method to stabilize shaky videos based on the contents of the frames and an adaptive blocking strategy. Each frame of the shaky video is adaptively divided into a triangle mesh according to the extracted feature trajectories. For different frames, the number and sizes of the triangles of the obtained meshes can be quite different. There are two reasons for this adaptive mesh operation.
\begin{itemize}
\item Parallax and discontinuous depth variation among different objects are more likely to happen in detail-rich regions. In these regions, mesh triangles should be small so that a single transformation matrix inside each triangle can well handle the contents. Fortunately, in these regions, more feature trajectories can usually be extracted and produce more triangles, i.e., a finer mesh, which is beneficial for more precise jitter estimation and warping.
\item As shown in Fig. \ref{fig:description}, we divide frames into triangle meshes according to the distribution of feature trajectories. These meshes have strong inter-frame correlation, i.e., the triangle meshes of neighboring frames are similar and vary slowly. This fact can help keep the local continuity of the adjacent stabilized frames and avoid sudden changes.
\end{itemize}
Based on the generated adaptive meshes,  meshed transformations are calculated by solving a two-stage optimization problem. To further enhance the robustness of the obtained meshed transformations, two adaptive weighting mechanisms are proposed to improve the spatial and temporal adaptability.

The remainder of this paper is organized as follows. Section \ref{sec:related_work} presents the related work regarding video stabilization. Section \ref{sec:method} presents the details of our adaptively meshed video stabilization method and two adaptive weighting mechanisms. Section \ref{sec:experiments} compares our method with some start-of-the-art stabilization methods through several public videos. Some concluding remarks are placed in Section \ref{sec:conclusion}.

\section{Related Work}
\label{sec:related_work}
\subsection{Traditional Video Stabilization}
Traditional video stabilization methods can be classified into three types, including 2D, 3D and 2.5D methods. 2D methods aim to stabilize videos mainly containing planar motion. Through image matching technologies, such as feature matching, 2D methods usually model the camera motion with inter-frame transformation matrix sequences \cite{chen2008capturing}, \cite{gleicher2008re}. Under the assumption of static scenes, many smoothing techniques are implemented to smooth the obtained transformation matrices, such as Gaussian low-pass filtering \cite{matsushita2006full}, Particle filtering \cite{yang2009robust} and  Regularization \cite{chang2004robust}, then stabilized frames are generated with the smoothed transform matrices. Grundmann {\it et al.} \cite{grundmann2011auto} proposed a linear programming framework to calculate a global homography through minimizing the first, second and third derivatives of the resulting camera path.  \cite{grundmann2012calibration} extended the method of \cite{grundmann2011auto} by replacing the global homography with a homography array  to reduce the rolling shutter distortion. Joshi {\it et al.} proposed a hyperlapse method \cite{joshi2015real}, which optimally selects some frames from the input frames, ensures that the selected frames can best match a desired speed-up rate and also result in the smoothest possible camera motion, and stabilizes the video according to these selected frames. Meanwhile, Zhang {\it et al.} \cite{zhang2018intrinsic} introduced the {\it ``as-similar-as-possible''} constraint to make the motion estimation more robust. 2D methods are computationally efficient and robust against planar camera motions, but may fail in the existence of strong parallax.

Parallax can be well handled by 3D methods. By detecting a set of feature points and tracking them at each frame, 3D methods reconstruct the 3D locations of these feature points and the 3D camera motion by the Structure-from-Motion (SFM) technique \cite{hartley2003multiple}. Thus these 3D methods are more complicated than 2D methods. Buehler {\it et al.} \cite{buehler2001non} got the smoothed locations of feature points by limiting the speed of the projected feature points to be constant. Liu {\it et al.} \cite{liu2009content} developed a 3D content-preserving warping technique. In \cite{liu2012video}, video stabilization is solved with an additional depth sensor, such as a Kinect camera. Moreover, some plane constraints were introduced to improve video stabilization performance \cite{wang2013multiplane}, \cite{zhou2013plane}. Due to the implementation of SFM, 3D methods usually stabilize videos much slower than 2D methods and may be fragile in the case of planar motions.

2.5D methods combine the advantages of 2D methods and 3D methods. Liu {\it et al.} \cite{liu2011subspace} extracted robust feature trajectories from the input frames and smoothed the camera path with subspace constraints. They also handled parallax by the content preserving warping \cite{wang2013spatially}. As this subspace property may not hold for dynamic scenes where cameras move quickly, Goldstein {\it et al.} introduced  epipolar constraints in \cite{goldstein2012video} and employed time-view reprojection for foreground feature trajectories.  In \cite{liu2014steadyflow}, Liu {\it et al.} designed a specific optical flow by enforcing strong spatial coherence and further extended it to meshflow \cite{liu2016meshflow} for sparse motion representation. However, these methods may fail when large foreground objects exist. So Ling {\it et al.} \cite{ling2018stabilization} took both foreground and background feature trajectories to estimate the camera jitters through solving an optimization problem. In \cite{zhao2019robust}, Zhao {\it et al.} implemented fast video stabilization assisted by foreground feature trajectories. Kon {\it et al.} \cite{koh2015video} generated virtual trajectories by augmenting incomplete trajectories using a low-rank matrix completion scheme when the number of the estimated trajectories is insufficient. In \cite{ma2019effective}, a novel method combines trajectory smoothing and frame warping into a single optimization framework and can conveniently increase the strength of the trajectory smoothing.

\subsection{CNN-based Video Stabilization}
Recently convolutional neural networks(CNNs) were implemented for video stabilization. Wang {\it et al.} \cite{wang2018deep} built a novel video stabilization dataset, called {\it DeepStab} dataset which includes 61 pairs of stable and unstable videos, and designed a network for multi-grid warping transformation learning, which can achieve comparable performance as traditional methods in regular videos and is more robust for low-quality videos. In \cite{xu2018deep}, adversarial networks are taken to estimate an affine matrix, with which steady frames can be generated. Huang {\it et al.} \cite{huang2019stablenet} designed a novel network which processes each unsteady frame progressively in a multi-scale manner from low resolution to high resolution, and then outputs an affine transformation to stabilize the frame. In \cite{choi2019deep}, a framework with frame interpolation techniques is utilized to generate stabilized frames. Generally speaking, CNN-based methods have great potential to robustly stabilize various videos, but may be limited by the lack of training video datasets.

\section{Proposed Method}
\label{sec:method}
As a 2.5D method, our method estimates the camera motion, smooths it and generates stabilized frames based on feature trajectories.
Compared with previous 2.5D methods, which handle parallax and discontinuous depth variation caused by foreground objects with fixed meshes, our method adaptively adjusts meshes by an adaptive blocking strategy and works well against different complicated scenes.
In the following subsections, we first introduce the feature trajectory pre-processing and related mathematical definitions, then will present the key steps of the proposed method.

\subsection{Pre-processing and mathematical definitions}
\label{pre-processing}

We adopt the KLT tracker of \cite{shi1994good} to detect feature points at each frame and track them to generate feature trajectories. To avoid all feature trajectories gathering in some small regions, we first divide each video frame into 10 by 10 grids and detect 200 corners with a global threshold. For grids which have no detected corners, we will decrease their local thresholds to ensure that at least one corner can be detected in each grid. Then these corners are tracked to generate feature trajectories. Thus there will be more feature trajectories in the grids with rich contents while less corners in grids with poor contents. As \cite{ma2019effective},  the corner detection procedure is executed only when the number of tracked feature points decreases to a certain level. However, in our method, much fewer corners are detected (1000 corners in \cite{ma2019effective}) since we directly produce the triangle mesh according to the positions of feature trajectories in each frame. We keep all trajectories whose length is longer than 3 frames.

\begin{figure}
   \centering
   \subfigure[Original image.]{
          \includegraphics[width=2.5in]{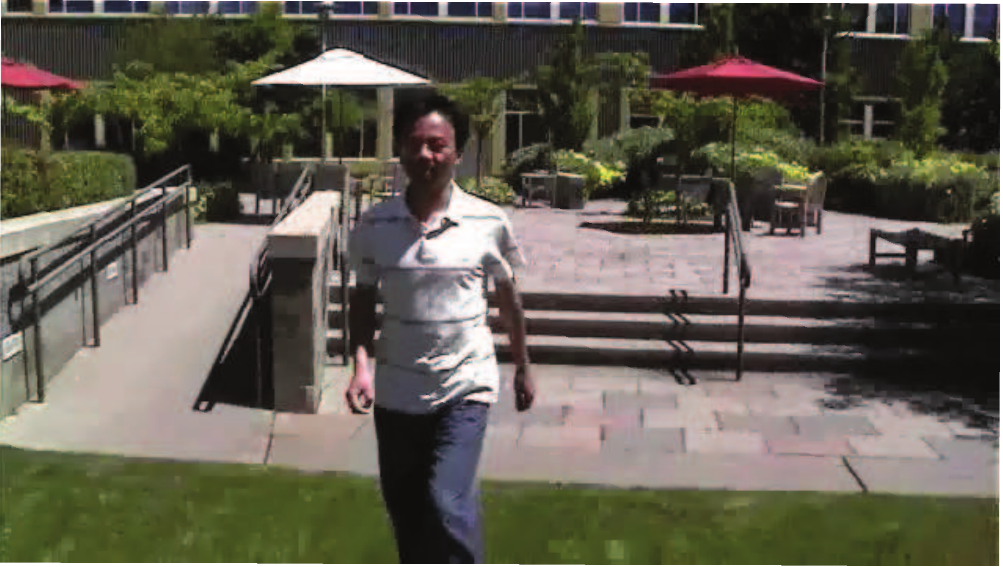}}
   \subfigure[The triangle mesh generated by feature trajectories.]{
          \includegraphics[width=2.5in]{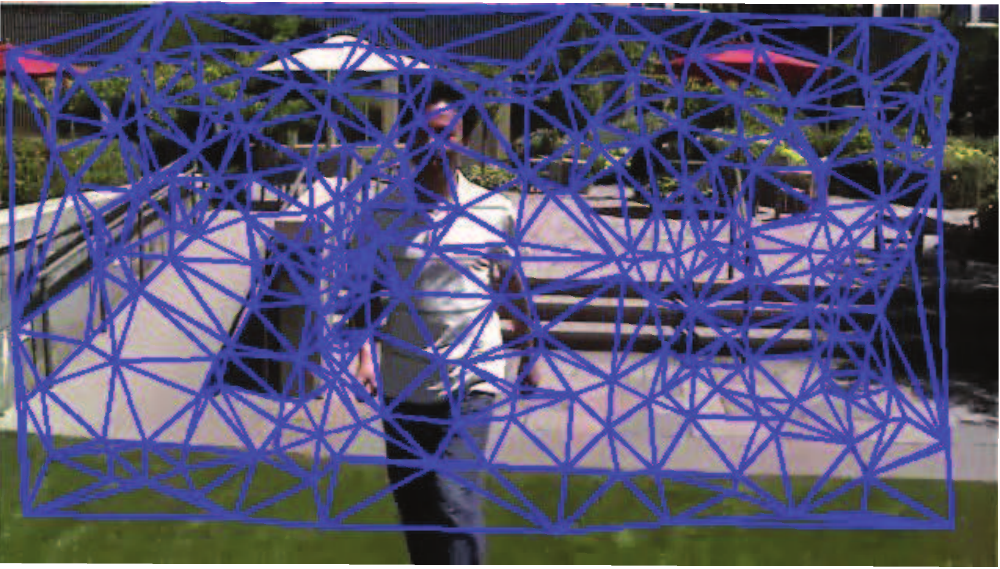}}
    \subfigure[The triangle mesh generated by feature trajectories and control points.]{
          \includegraphics[width=2.5in]{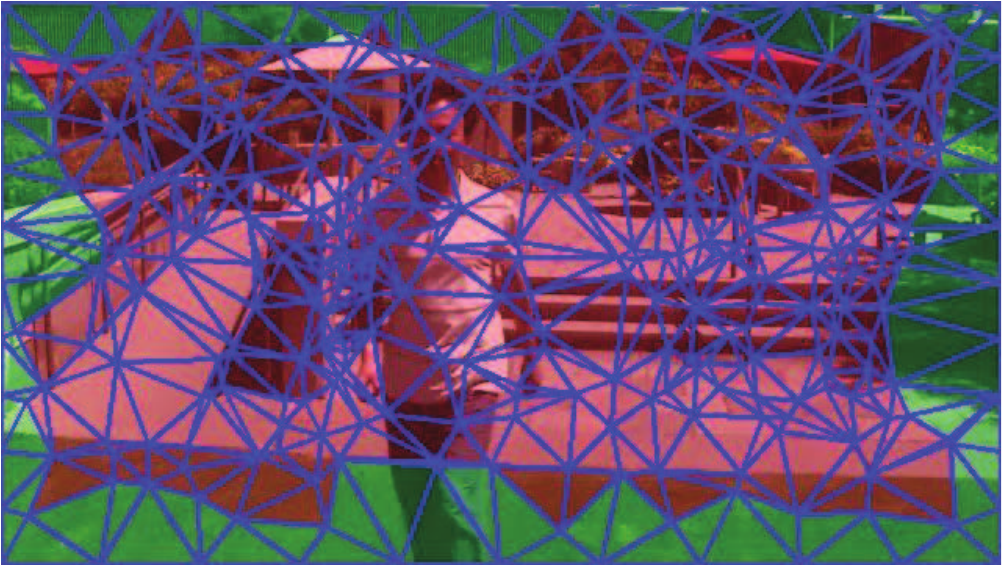}}
   \caption{Triangle meshes of an example frame through standard constrained Delaunay triangulation.}
\label{fig:Delaunay}
\end{figure}

Suppose there are $N$ feature trajectories in a shaky video, which are denoted as $\{P_i\}_{i=1}^N$. These feature trajectories may come from either the background or foreground objects. Denote the first and last frames, in which trajectory $i$($i\in [1,N]$) appears, as $s_i$ and $e_i$. Denote the feature point of trajectory $i$ at frame $t$ ($s_i\leq t\leq e_i$) as $P_{i,t}$. The feature points of all feature trajectories which appear at frame $t$ are grouped into a set,
\begin{eqnarray}
\label{eq:P_to_M}
M_t=\{P_{i,t}|i \in [1,N], t \in[s_i,e_i] \}.
\end{eqnarray}
 Then we generate a triangle mesh of frame $t$ with $M_t$ through standard constrained Delaunay triangulation\cite{igarashi2005rigid}, which is illustrated in Fig. \ref{fig:Delaunay}(b). Suppose there are $K_t$ triangles in the triangle mesh of frame $t$, then the set of triangles of frame $t$ is denoted as $Q_t=\{Q_{t,1},Q_{t,2},...,Q_{t,K_t}\}$. The stabilized views of $M_t$ and $Q_t$ are represented as $\widetilde{M}_t$ and $\widetilde{Q}_t$, respectively.

Our method estimates $\{\widetilde{P}_i\}_{i=1}^N$, i.e., the stabilized views of feature trajectories $\{P_i\}_{i=1}^N$, through solving an optimization problem under the following three constraints.
\begin{enumerate}
\item For a given feature trajectory $P_i$, its stabilized view $\widetilde{P}_i$ should vary slowly.
\item At frame $t$, each stabilized triangle $\widetilde{Q}_t$ is geometrically similar to its original triangle $Q_t$.
\item At frame $t$, the transformations between $\widetilde{Q}_t$ and its adjacent triangles  should be similar to the corresponding transformations between $Q_t$ and its adjacent triangles.
\end{enumerate}

As shown in Fig. \ref{fig:Delaunay}(b),  all triangles are constructed with feature trajectories and  we can calculate their stabilized views and warp these regions according to the corresponding transformations.
%Some boundary regions are left without being stabilized.
At each frame, our method also evenly sets $10$(other number should also be fine) control points on each of four frame edges and produces $36$ control points which are denoted as $E_t=\{E_{t,1},...,E_{t,36}\}$. Then we divide frame $t$ with the new set $\{ M_t, E_t\}$ through Delaunay triangulation to produce a triangle mesh. The obtained triangle mesh is represented as $\{Q_{t,1},...,Q_{t,K_t},B_{t,1},...,B_{t,L_t}\}$, where $Q_t=\{Q_{t,1},...,Q_{t,K_t}\}$, called {\it inner triangles}, are $K_t$ triangles whose vertices are made up of only $M_t$, and $B_t=\{B_{t,1},...,B_{t,L_t}\}$, called {\it outer triangles}, are $L_t$ triangles whose vertices contain at least one control point. In Fig. \ref{fig:Delaunay}(c), {\it inner triangles} are presented in red and {\it outer triangles} in green. Then a two-stage optimization problem is solved to calculate $\widetilde{M_t}$ and $\widetilde{E}_t$, i.e., the stabilized views of $M_t$ and $E_t$. At the first stage, a global optimization problem is solved to estimate the stabilized views of feature trajectories $\{P_i\}_{i=1}^N$, i.e., $\{\widetilde{P}_i\}_{i=1}^N$, which form $\widetilde{M}_t$ as (\ref{eq:P_to_M}).  At the second stage, a frame-by-frame optimization problem is solved to produce $\widetilde{E}_t$. Finally, each frame is warped through the transformations between the triangles generated by $\{ \widetilde{M}_t, \widetilde{E}_t\}$ and the ones by $\{M_t, E_t\}$.

\subsection{Stabilized view estimation from feature trajectories $\{P_i\}_{i=1}^N$}
\label{sec:estimate_feature}
Based on the three constraints in Section \ref{pre-processing}, the stabilized view of $\{P_i\}_{i=1}^N$ is obtained through solving the following optimization problem,

\begin{eqnarray}
\label{def:optim}
\min_{\{\widetilde{P}_i\}_{i=1}^N} \mathcal{O}(\{\widetilde{P}_i\}_{i=1}^N),
\end{eqnarray}
where
\begin{eqnarray}
\label{eq:optimization}
%\small
&&\mathcal{O}( \{\widetilde{P}_i\}_{i=1}^N)  = \\
&&\mathcal{O}_{Smooth}+\mathcal{O}_{InterSim}+\mathcal{O}_{IntraSim}+\mathcal{O}_{Reg},\nonumber
\end{eqnarray}
$\mathcal{O}_{Smooth}$ is the smoothing term to smooth feature trajectories, $\mathcal{O}_{InterSim}$ is the inter-frame similarity transformation term to ensure that the warped frame closely follows the similarity transformation of its original frame, $\mathcal{O}_{IntraSim}$ is the intra-frame similarity transformation term to ensure that the similarity transformation between adjacent regions of a stabilized frame is close to the corresponding transformation of its original frame, and $\mathcal{O}_{Reg}$ is the regularization term to avoid too much warping. The details of these four terms are provided below.
%\begin{eqnarray*}
%\mathcal{O}_1 &=& \sum_{i=1}^N\sum_{j=s_i+1}^{e_i}\alpha_{i,j}\|\widetilde{P}_{i,j}-\widetilde{P}_{i,j-1}\|^2, \\
%\mathcal{O}_2&=&\sum_{i=1}^N\sum_{j=s_i+1}^{e_i-1}\beta_{i,j}\|(\widetilde{P}_{i,j+1}-\widetilde{P}_{i,j})-(\widetilde{P}_{i,j}-\widetilde{P}_{i,j-1})\|^2,\\
%\mathcal{O}_3 &=& \sum_{t=1}^T\sum_{i=1}^{N_Q}\sum_{j={1,2,3}}\gamma\|\widehat{Q}_{t,i}^j-\widetilde{Q}_{t,i}^j\|^2, \\
%\mathcal{O}_4 &=& \sum_{t=1}^T\sum_{i=1}^{N_Q}\sum_{j\in adj(i)}\varepsilon\|\overline{Q}_{t,i}^j-\overline{Q}_{t,i}^j\|^2, \\
%\mathcal{O}_5 &=& \sum_{i=0}^N\sum_{j=s_i}^{e_i} \|(\widetilde{P}_{i,j}-P_{i,j})\|^2,
%\end{eqnarray*}

\subsubsection{The smoothing term $\mathcal{O}_{smooth}$}
$ $

Motivated by \cite{grundmann2011auto}, we smooth the feature trajectories through their first-order and second-order derivatives. The first-order derivative term requires each stabilized feature trajectory $P_i$ to change slowly, i.e., we expect
\begin{equation}
\|P_{i,k}-P_{i,k-1}\|^2=0, k \in [s_i+1,e_i].
\end{equation}

If this requirement is satisfied, the camera will be static and the generated video will be the most stable. However, that extreme requirement may cause much cropping and discard a lot of contents of the video. So we introduce the second-derivative term, which requires the velocity of each feature trajectory $P_i$ to be constant, i.e.,
\begin{equation}
\label{eq:second_derivative}
\|(P_{i,k+1}-P_{i,k})-(P_{i,k}-P_{i,k-1})\|^2=0, k \in [s_i+1,e_i-1].
\end{equation}
The term in (\ref{eq:second_derivative}) will limit the change of the inter-frame motion and produce stable feature trajectories.  Combining the first-order derivative term and the second-order derivative term, we define $\mathcal{O}_{Smooth}$ as

\begin{eqnarray}
\mathcal{O}_{Smooth}&=&\mathcal{O}_{Smooth1}+\mathcal{O}_{Smooth2} \nonumber \\
&=&\sum_{i=1}^N\sum_{j=s_i+1}^{e_i}\alpha\|\widetilde{P}_{i,j}-\widetilde{P}_{i,j-1}\|^2 \nonumber \\
&+&\sum_{i=1}^N\sum_{j=s_i+1}^{e_i-1}\beta\|(\widetilde{P}_{i,j+1}-\widetilde{P}_{i,j})-(\widetilde{P}_{i,j}-\widetilde{P}_{i,j-1})\|^2,\nonumber\\
\label{eq:smooth}
\end{eqnarray}
where $\alpha$ and $\beta$ are weighting parameters to be determined later.

\subsubsection{The inter-frame similarity transformation term $\mathcal{O}_{InterSim}$}
$ $

\begin{figure}[!hbt]
\centering\includegraphics[width=\hsize]{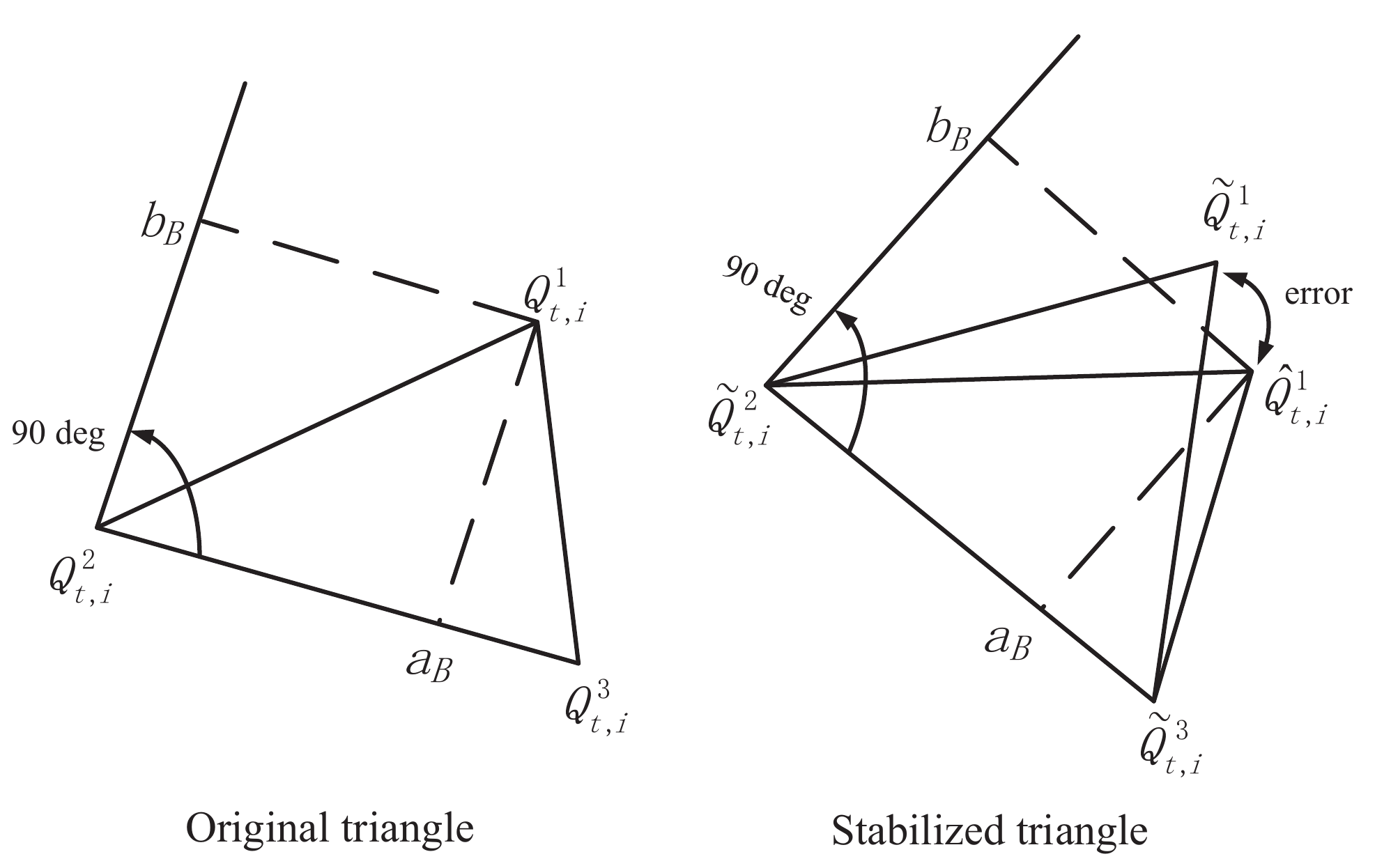}
\caption{A triangle vertex can be locally expressed/calculated by the other two vertices. The deviation from a similarity transformation of a warp can be measured as the distance between the desired vertex and the calculated vertex.}
\label{fig:inter_similarity}
\end{figure}

This term is designed by following the {\it ``as-similar-as-possible''} warping method of \cite{liu2009content}. For $K_t$ Delaunay triangles in $Q_t$, their stabilized views are denoted as  $\widetilde{Q}_t=\{\widetilde{Q}_{t,1},\widetilde{Q}_{t,2},...,\widetilde{Q}_{t,K_t}\}$. The vertices of triangle $i$ in $Q_t$ and $\widetilde{Q}_t$ are denoted as $\{Q_{t,i}^1,Q_{t,i}^2,Q_{t,i}^3\}$ and $\{\widetilde{Q}_{t,i}^1,\widetilde{Q}_{t,i}^2,\widetilde{Q}_{t,i}^3\}$, respectively. To preserve the local shape of frame $t$,  $\widetilde{Q}_t$ is required to be geometrically similar to $Q_t$. As shown in Fig. \ref{fig:inter_similarity}, for triangle $i$ in $Q_t$, any vertex can be locally represented by the other two vertices. For example, we have
\begin{equation}
Q_{t,i}^1=Q_{t,i}^2+a_B(Q_{t,i}^3-Q_{t,i}^2)+b_BR_{90}(Q_{t,i}^3-Q_{t,i}^2),
\end{equation}
where $R_{90}=[0,1;-1,0]$, and $a_B$ and $b_B$ are the coordinates of $Q_{t,i}^1$ in the local coordinate system defined by $Q_{t,i}^2$ and $Q_{t,i}^3$.
Based on the geometric similarity constraint between  $Q_{t,i}$ and $\widetilde{Q}_{t,i}$, the expected position of $\widetilde{Q}_{t,i}^1$ is

\begin{equation}
\widehat{Q}_{t,i}^1=\widetilde{Q}_{t,i}^2+a_B(\widetilde{Q}_{t,i}^3-\widetilde{Q}_{t,i}^2)+b_BR_{90}(\widetilde{Q}_{t,i}^3-\widetilde{Q}_{t,i}^2).
\end{equation}

Our goal is to minimize $\|\widetilde{Q}_{t,i}^1-\widehat{Q}_{t,i}^1\|^2$. Similar constraints are performed for $\widetilde{Q}_{t,i}^2$ and $\widetilde{Q}_{t,i}^3$ and the overall
inter-frame similarity transformation term is defined as
\begin{equation}
\mathcal{O}_{InterSim} = \sum_{t=1}^T\sum_{i=1}^{K_t}\sum_{j=\{1,2,3\}}\gamma\|\widehat{Q}_{t,i}^j-\widetilde{Q}_{t,i}^j\|^2, \\
\end{equation}
where $T$ is the total number of frames of the concerned video. Note that since $Q_t$ is made up of $M_t$, the constraints for $\widetilde{Q}_t$ is equivalent to the constraints for $\widetilde{M}_t$, which is actually the set of feature points of stabilized trajectories $\{\widetilde{P}_i\}_{i=1}^N$ at each frame.

\subsubsection{The intra-frame similarity transformation term $\mathcal{O}_{IntraSim}$}
\label{sec:estimate_feature_intra}
$ $

\begin{figure}[!hbt]
\centering\includegraphics[width=\hsize]{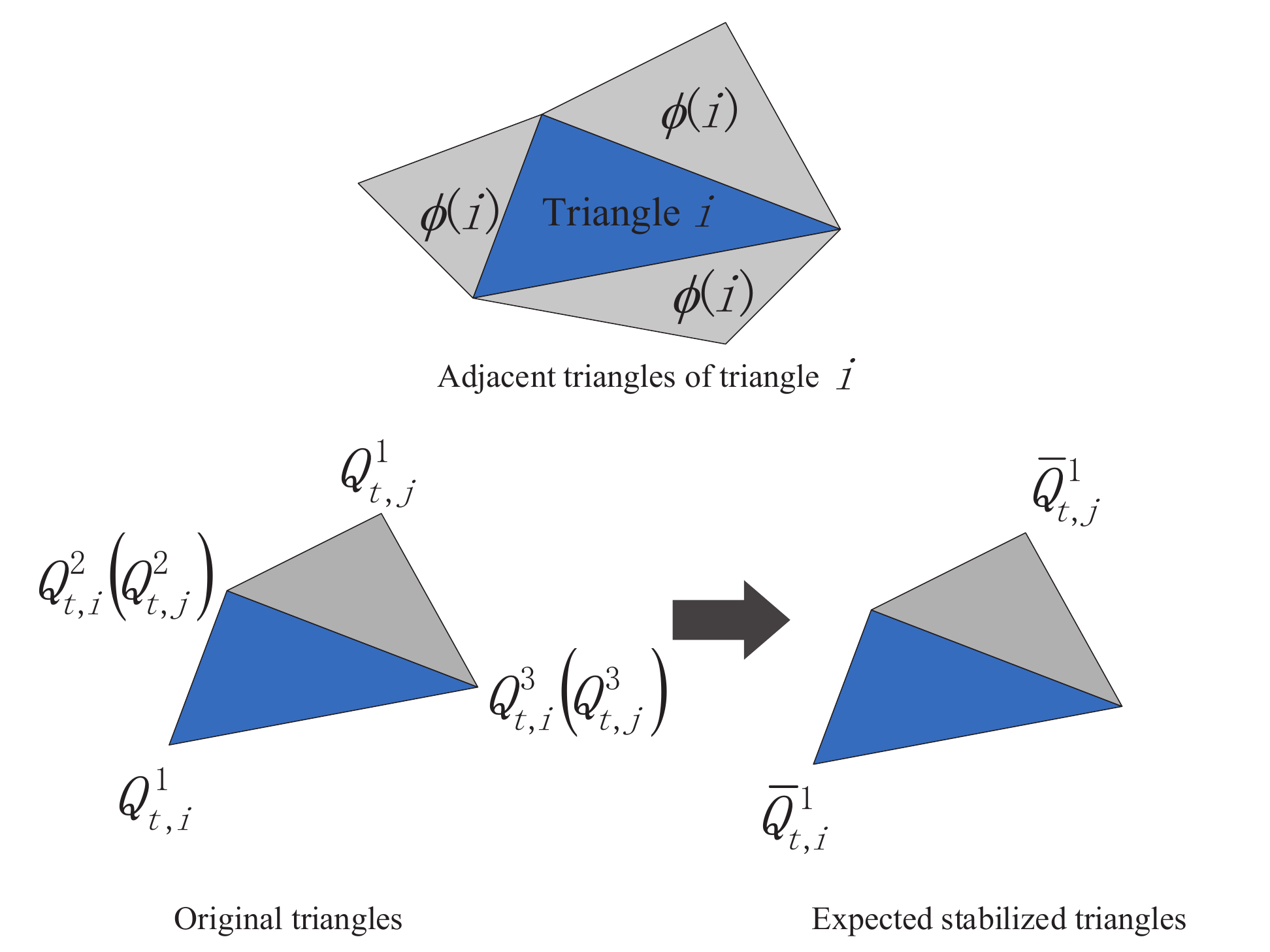}
\caption{Illustration of the intra-frame similarity transformation term. The first row shows triangle $i$ and its adjacent triangles which comprise the set $\phi(i)$. The second row shows triangle $i$ and one of its adjacent triangles, triangle $j$. We first represent $Q_{t,i}^1$ and $Q_{t,j}^1$  with the weights of the other triangle and calculate $\overline{Q}_{t,i}^j$ and $\overline{Q}_{t,j}^i$ with the corresponding weights.
%The deviation of the transformation between adjacent triangles in the shaky frame and the its stabilized view are measured by $\|\overline{Q}_{t,i}^1-\widetilde{Q}_{t,i}^1\|^2$ and $\|\overline{Q}_{t,j}^1-\widetilde{Q}_{t,j}^1\|^2$
}
\label{fig:intra_sim}
\end{figure}

To avoid distortion in stabilized frames, this term requires that the transformations between adjacent regions in each stabilized frame should be closely followed by those corresponding transformations in its original frame. As shown in Fig. \ref{fig:intra_sim}, in the triangle mesh, each triangle and its adjacent triangles share one common edge. For triangle $i$ at frame $t$, we take one of its neighbors, triangle $j$, as an example. Triangles $i$ and $j$ share two vertices which are denoted as $Q_{t,i}^2(Q_{t,j}^2)$ and $Q_{t,i}^3(Q_{t,j}^3)$, while the other vertices of the two triangles are $Q_{t,i}^1$ and $Q_{t,j}^1$, respectively. Then $Q_{t,i}^1$ can be represented by the coordinates of triangle $j$ through the standard linear texture mapping \cite{igarashi2005rigid},
%: $Q_{t,i}^1=aQ_{t,j}^1+bQ_{t,j}^2+cQ_{t,j}^3$, and $a$, $b$ and $c$ are calculated by,
\begin{eqnarray}
\left\{\begin{array}{ll}
Q_{t,i}^1(x)=aQ_{t,j}^1(x)+bQ_{t,j}^2(x)+cQ_{t,j}^3(x)\\
Q_{t,i}^1(y)=aQ_{t,j}^1(y)+bQ_{t,j}^2(y)+cQ_{t,j}^3(y)\\
a+b+c=1
\end{array}\right.,
\label{eq:linear_texture_mapping}
\end{eqnarray}
where $Q_{t,i}^1(x)$ and $Q_{t,i}^1(y)$ are the horizontal and vertical components of $Q_{t,i}^1$, $a$, $b$ and $c$ are weighting parameters.  With these weights, $\widetilde{Q}_{t,j}^1$, $\widetilde{Q}_{t,j}^2$ and $\widetilde{Q}_{t,j}^3$, the expected  position of  $\widetilde{Q}_{t,i}^1$ is calculated as
\begin{equation}
\overline{Q}_{t,i}^1 = a\widetilde{Q}_{t,j}^1+b\widetilde{Q}_{t,j}^2+c\widetilde{Q}_{t,j}^3.
\end{equation}

In order to ensure the transformation between $\widetilde{Q}_{t,i}$ and $\widetilde{Q}_{t,j}$ is close to the one between $Q_{t,i}$ and $Q_{t,j}$,  we want to minimize the  difference between  $\overline{Q}_{t,i}^1$ and its stabilized coordinate $\widetilde{Q}_{t,i}^1$, i.e.,
\begin{equation}
\|\overline{Q}_{t,i}^1-\widetilde{Q}_{t,i}^1\|^2=0. \\
\end{equation}

Similarly, we can calculate $\overline{Q}_{t,j}^1$ and  reduce the difference between $\overline{Q}_{t,j}^1$ and $\widetilde{Q}_{t,j}^1$. The total intra-frame similarity transformation term is defined as

\begin{equation}
\label{def:intrasim}
\mathcal{O}_{IntraSim} = \sum_{t=1}^T\sum_{i=1}^{K_t}\sum_{j\in \phi(i)}\varepsilon(\|\overline{Q}_{t,i}^1-\widetilde{Q}_{t,i}^1\|^2+\|\overline{Q}_{t,j}^1-\widetilde{Q}_{t,j}^1\|^2),
\end{equation}
where $\phi(i)$ is the set of adjacent triangles of triangle $i$, and $\varepsilon$ is a weighting parameter to be determined later.

\subsubsection{The regularization term $\mathcal{O}_{Reg}$}
$\mathcal{O}_{Reg}$ aims to regularize the optimization in (\ref{eq:optimization}) so that the stabilized trajectories stay close to their original ones to avoid too much warping. It is defined as
\begin{equation}
\mathcal{O}_{Reg} = \sum_{i=0}^N\sum_{j=s_i}^{e_i} \|(\widetilde{P}_{i,j}-P_{i,j})\|^2.
\end{equation}

With the above optimization in (\ref{eq:optimization}), we can get the stabilized views of all feature trajectories, i.e., $\widetilde{P}_{i,k}$ ($i\in [1,N], k \in[s_i,e_i]$).
%To enhance the robustness of our method, we will further introduce two adaptive weighting mechanisms to adjust the weighting parameters of the optimization . The parameter sensitivity will be analyzed through experiments.

\subsection{Stabilized view estimation from control points}

\begin{figure}
\centering\includegraphics[width=0.8\hsize]{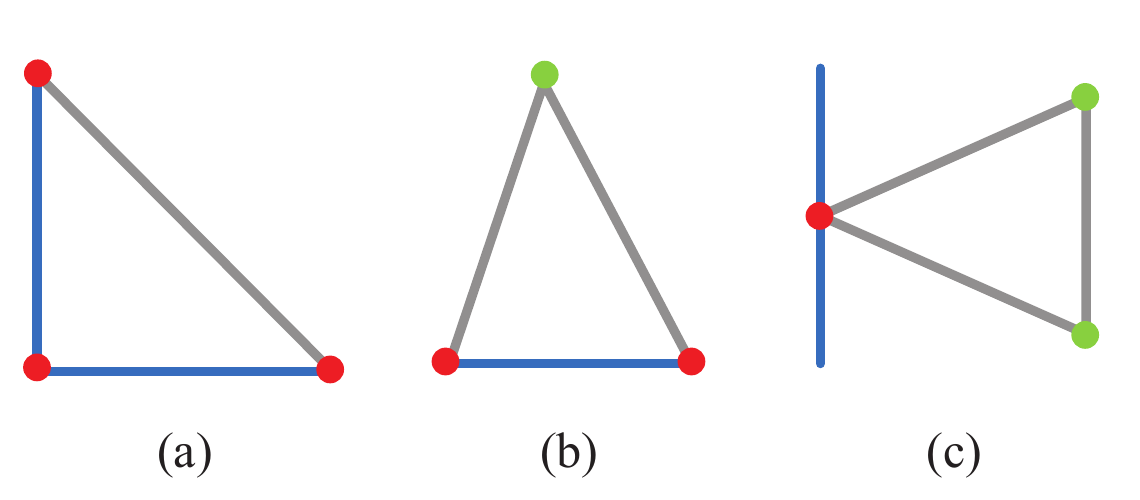}
\caption{Three types of triangles in $B_t$. The blue lines represent the edges of the video frame. The red points stand for the control points and the green points stand for the points of feature trajectories. (a) A triangle containing three control points. (b) A triangle containing two control points. (c) A triangle containing one control point.}
\label{fig:boundary_cases}
\end{figure}

Given the stabilized views of feature trajectories, $\{\widetilde{P}_i\}_{i=1}^N$, we have the stabilized views of $M_t$ and $Q_t$ at each frame, i.e., $\widetilde{M}_t$ and $\widetilde{Q}_t$. At each frame, we estimate the stabilized view of control points $E_t$, i.e., $\widetilde{E}_t$, by solving an optimization problem. Then the stabilized view of $B_t$, $\widetilde{B}_t=\{\widetilde{B}_{t,1},\widetilde{B}_{t,2},...,\widetilde{B}_{t,L_t}\}$,  can be computed from $\widetilde{M}_t$ and $\widetilde{E}_t$.  Finally, each frame is warped according to the transformations between $\{Q_t, B_t\}$ and $\{\widetilde{Q}_{t}, \widetilde{B}_t\}$. There are three types of triangles in $B_t$, triangles with $1$, $2$ or $3$ control points, which are shown in Fig. \ref{fig:boundary_cases}.  We design an optimization problem to  calculate only the stabilized views of control points in $B_t$ as follows,

\begin{eqnarray}
\label{def:optim_boundary}
\min_{\{\widetilde{B}_t\}} \mathcal{O}(\{\widetilde{B}_t\}),
\end{eqnarray}
where
\begin{eqnarray}
\mathcal{O}( \{\widetilde{B}_t\})  = \mathcal{O}_{InterSim}+\mathcal{O}_{IntraSim},
\label{eq:optimization_boundary}
\end{eqnarray}
 and the definitions of $\mathcal{O}_{InterSim}$ and $\mathcal{O}_{IntraSim}$ are given below.

\subsubsection{The inter-frame similarity transformation term $\mathcal{O}_{InterSim}$}
$ $

This term is also designed to ensure that  $\widetilde{B}_t$ is geometrically similar to $B_t$. For triangle $i$ in $B_t$, we have
\begin{equation}
B_{t,i}^1=B_{t,i}^2+a_B(B_{t,i}^3-B_{t,i}^2)+b_B R_{90}(B_{t,i}^3-B_{t,i}^2),
\end{equation}
where $R_{90}=[0,1;-1,0]$, and $a_B$ and $b_B$ are the coordinates of $B_{t,i}^1$ in the local coordinate system defined by $B_{t,i}^2$ and $B_{t,i}^3$.
Based on the geometric similarity constraint between  $B_{t,i}$ and $\widetilde{B}_{t,i}$,  the expected position of $\widetilde{B}_{t,i}^1$ is

\begin{equation}
\widehat{B}_{t,i}^1=\widetilde{B}_{t,i}^2+a_B(\widetilde{B}_{t,i}^3-\widetilde{B}_{t,i}^2)+b_B R_{90}(\widetilde{B}_{t,i}^3-\widetilde{B}_{t,i}^2).
\label{eq:inter_boundary}
\end{equation}
Here $\widetilde{B}_{t,i}$ may contain one or two vertices, which also belong to triangles of $\widetilde{M}_t$. These common vertices have been calculated by (\ref{def:optim}) and do not need to be optimized again. Therefore, in (\ref{eq:inter_boundary}), we replace these common vertices of $\widetilde{B}_{t,i}$, denoted as $\widetilde{B}_{t,i}^j$, with the corresponding stabilized vertices in triangles of $\widetilde{M}_t$, which are denoted as $\Psi_{\widetilde{M}_t}(\widetilde{B}_{t,i}^j)$.
So the inter-frame similarity transformation term is defined as

\begin{equation}
\mathcal{O}_{InterSim} =\sum_{i=1}^{L_t}\sum_{j=\{1,2,3\}}\gamma \Upsilon (\widetilde{B}_{t,i}^j), \\
\end{equation}
where $\gamma$ is a weighting parameter to be determined later, and $\Upsilon(\cdot)$ is defined as
\begin{eqnarray}
\Upsilon(\widetilde{B}_{t,i}^j) =
\left\{\begin{array}{ll}
\|\widehat{B}_{t,i}^j - \widetilde{B}_{t,i}^j\|^2, &\widetilde{B}_{t,i}^j \in \widetilde{E}_t \\
\|\widehat{B}_{t,i}^j - \Psi_{\widetilde{M}_t}(\widetilde{B}_{t,i}^j)\|^2, &\widetilde{B}_{t,i}^j \in \widetilde{M}_t
\end{array}\right..\nonumber\\
\label{eq:betai_gamma}
\end{eqnarray}

\subsubsection{The intra-frame similarity transformation term $\mathcal{O}_{IntraSim}$}
$ $

This term is similar to $\mathcal{O}_{IntraSim}$ in Section \ref{sec:estimate_feature_intra}. Transformations between adjacent triangles in each stabilized frame are also expected to be close to the corresponding transformations between adjacent triangles in its original frame. Note that the adjacent triangle $j$ of triangle $i$ may come from $B_t$ or $Q_t$.
% and some vertices of $B_{t,i}$ may come from $M_t$ or $E_t$.
We simply combine $B_t$ and $Q_t$ into a larger set, which is denoted as $BQ_t=\{BQ_{t,1},BQ_{t,2},...,BQ_{t,K_t+L_t}\}$, and the adjacent triangle $j$ of $B_{t,i}$ is denoted as $BQ_{t,j}$ at frame $t$. The different vertices of $B_{t,i}$ and $BQ_{t,j}$ are denoted as $B_{t,i}^1$ and $BQ_{t,j}^1$.
Then $B_{t,i}^1$ can be represented by the coordinates of triangle $BQ_{t,j}$ through the linear texture mapping
\begin{eqnarray}
B_{t,i}^1=aBQ_{t,j}^1+bBQ_{t,j}^2+cBQ_{t,j}^3,
\end{eqnarray}
 where $a$, $b$ and $c$ can be similarly calculated as (\ref{eq:linear_texture_mapping}).
The expected coordinate $\overline{B}_{t,i}^1$ with these weights and corresponding vertices of the stabilized triangle $\widetilde{BQ}_{t,j}$, i.e., $\widetilde{BQ}_{t,j}^1$, $\widetilde{BQ}_{t,j}^2$, $\widetilde{BQ}_{t,j}^3$, are calculated as
\begin{equation}
\overline{B}_{t,i}^1 = a\widetilde{BQ}_{t,j}^1+b\widetilde{BQ}_{t,j}^2+c\widetilde{BQ}_{t,j}^3. \\
\end{equation}

$\overline{B}_{t,i}^1$ is expected to stay close to the  stabilized coordinate $\widetilde{B}_{t,i}^1$. When  vertices of $\widetilde{BQ}_{t,j}$ and $\widetilde{B}_{t,i}^1$ belong to some triangles of $\widetilde{M}_t$, these common vertices are replaced with the stabilized vertices of the corresponding triangles of $\widetilde{M}_t$.

Similarly, we have ${BQ}_{t,j}^1 = a{B}_{t,i}^1+b{B}_{t,i}^2+c{B}_{t,i}^3$ and $\overline{BQ}_{t,j}^1 = a\widetilde{B}_{t,i}^1+b\widetilde{B}_{t,i}^2+c\widetilde{B}_{t,i}^3$.  $\overline{BQ}_{t,j}^1$ is expected to stay close to $\widetilde{BQ}_{t,j}^1$. Finally the intra-frame similarity transformation term is defined as

\begin{equation}
\mathcal{O}_{IntraSim} = \sum_{i=1}^{L_t}\sum_{j\in \phi(i)}\varepsilon (\Gamma(\widetilde{B}_{t,i}^1)+\Gamma(\widetilde{BQ}_{t,j}^1)),
\end{equation}
where $\varepsilon$ is the same weighting parameter in (\ref{def:intrasim}), $\Gamma(\widetilde{B}_{t,i}^1)$  is defined as
\begin{eqnarray}
\Gamma(\widetilde{B}_{t,i}^1) =
\left\{\begin{array}{ll}
\|\overline{B}_{t,i}^1-\widetilde{B}_{t,i}^1\|^2, &\widetilde{B}_{t,i}^1 \in \widetilde{E}_t \\
\|\overline{B}_{t,i}^1-\Psi_{\widetilde{M}_t}(\widetilde{B}_{t,i}^1)\|^2, &\widetilde{B}_{t,i}^1 \in \widetilde{M}_t
\end{array}\right.,\nonumber\\
\label{eq:betai_gamma_1}
\end{eqnarray}
 and $\Gamma(\widetilde{BQ}_{t,j}^1)$ is similarly defined as $\Gamma(\widetilde{B}_{t,i}^1)$.

\subsection{Adaptive Weighting Mechanisms}
\label{sec:adaptive}
To enhance the robustness of our method, we further introduce two adaptive weighting mechanisms to adjust the weighting parameters of the optimization in (\ref{eq:optimization}). They aim to improve the spatial and temporal adaptability of our method.

\subsubsection{A weighting mechanism to improve the temporal adaptability}
\label{sec:temporal}
$ $

\begin{figure*}
\centering\includegraphics[width=0.9\hsize]{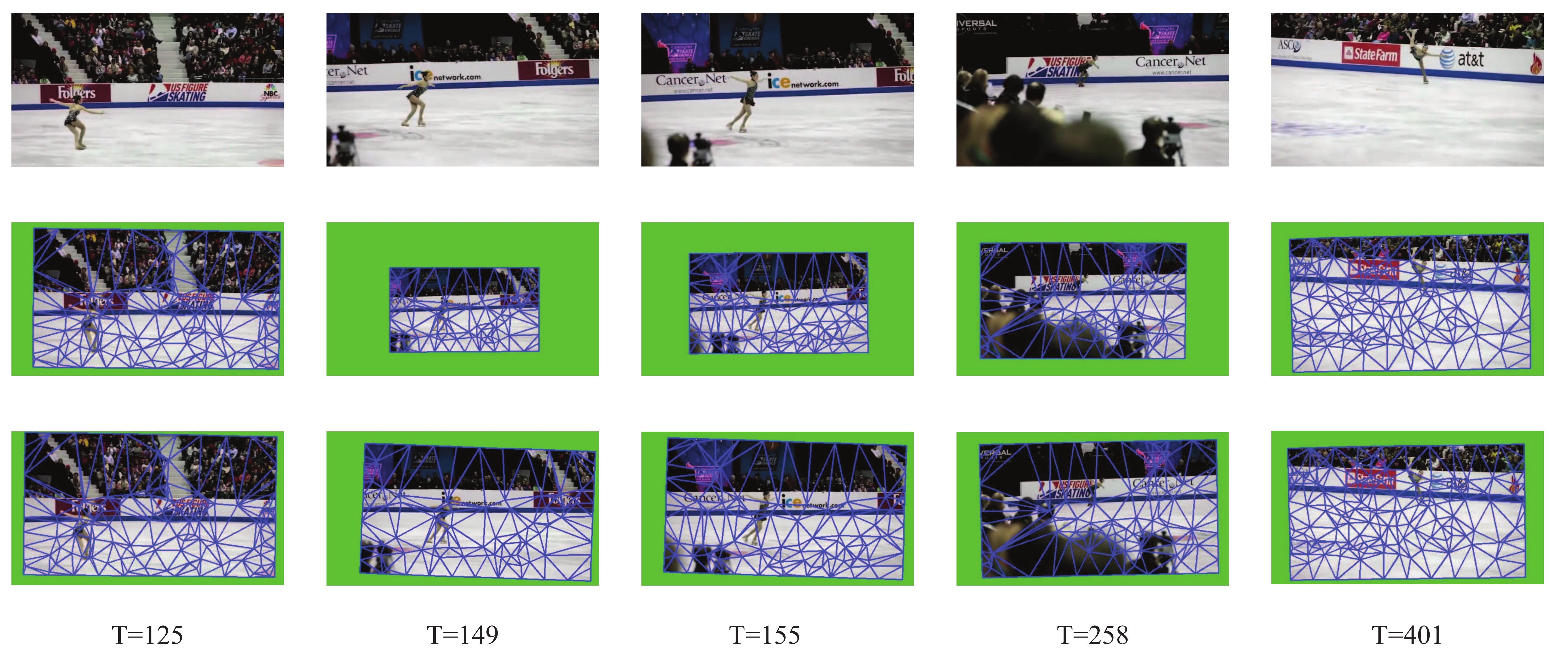}
\caption{The weighting mechanism for the temporal adaptability enables our method to handle videos with fast camera motion. The first row shows several original frames in a shaky video. The second row shows the corresponding stabilized frames with $\mathcal{O}_{Smooth1}$ in (\ref{eq:smooth}), where we can observe stabilization collapse. The third row shows the corresponding stabilized frames with improved $\mathcal{O}_{Smooth1}$ in (\ref{eq:improved_smooth1}), which well avoids the undesirable stabilization collapse.}
\label{fig:adaptive_temporal}
\end{figure*}

According to the analysis in \cite{ma2019effective}, $\mathcal{O}_{Smooth1}$ aims to reduce the inter-frame movement of a trajectory to zero. However, the fast camera motion usually produces huge inter-frame movement of feature trajectories.  $\mathcal{O}_{Smooth1}$ is enforced to reduce such inter-frame movement; otherwise, too big inter-frame movement may lead to the collapse of the stabilization of shaky videos. So the weight of the first-order derivative term in (\ref{eq:smooth}), i.e., $\alpha$, should be reduced when the camera moves fast. Different from the method of \cite{ma2019effective}, we consider the speed continuity of the camera motion, and update $\mathcal{O}_{Smooth1}$ into

\begin{eqnarray}
\mathcal{O}_{Smooth1}&=&\alpha \sum_{i=1}^N\sum_{j=s_i+1}^{e_i} (e^{-(\frac{\overline{d}_{x_i}^j-\sigma}{\sigma})^3} \|\widetilde{P}_{i,j}^x-\widetilde{P}_{i,j-1}^x\|^2 \nonumber \\
&+&e^{-(\frac{\overline{d}_{y_i}^j-\sigma}{\sigma})^3}\|\widetilde{P}_{i,j}^y-\widetilde{P}_{i,j-1}^y\|^2),
\label{eq:improved_smooth1}
\end{eqnarray}
where $\widetilde{P}_{i,j}^x$ and $\widetilde{P}_{i,j}^y$ are the horizontal and vertical components of $\widetilde{P}_{i,j}$, and ${P}_{i,j}^x$ and ${P}_{i,j}^y$ are the horizontal and vertical components of ${P}_{i,j}$. $\overline{d}_{x_i}^j$ and $\overline{d}_{y_i}^j$ measure the inter-frame movement of feature trajectory $P_i$ at frame $j$ and are calculated as $\overline{d}_{x_i}^j=P_{i,j}^x-P_{i,j-1}^x$ and $\overline{d}_{y_i}^j=P_{i,j}^y-P_{i,j-1}^y$.
$\sigma=10$ is a constant. When the camera moves fast, $\alpha$ decreases so that the smoothing strength of  $\mathcal{O}_{Smooth1}$ is reduced. This adaptive weighting mechanism enables our method to handle videos with fast camera motion. An example in Fig. \ref{fig:adaptive_temporal} shows the effects of this weighting mechanism on the temporal adaptability.

\subsubsection{A weighting mechanism to improve the spatial adaptability}
$ $

During stabilizing shaky videos, we cannot ignore discontinuous depth variation or different local motion caused by foreground objects; otherwise, significant distortion may appear in stabilized frames. To solve this problem, we first determine whether each triangle of the triangle mesh contains foreground objects or not, and then increase the weights of triangles containing foreground objects inside $\mathcal{O}_{InterSim}$ of (\ref{eq:optimization}). This operation will avoid serious distortion artifacts in complicated scenes and enhance the robustness of our optimization model.

\begin{figure}
\centering\includegraphics[width=\hsize]{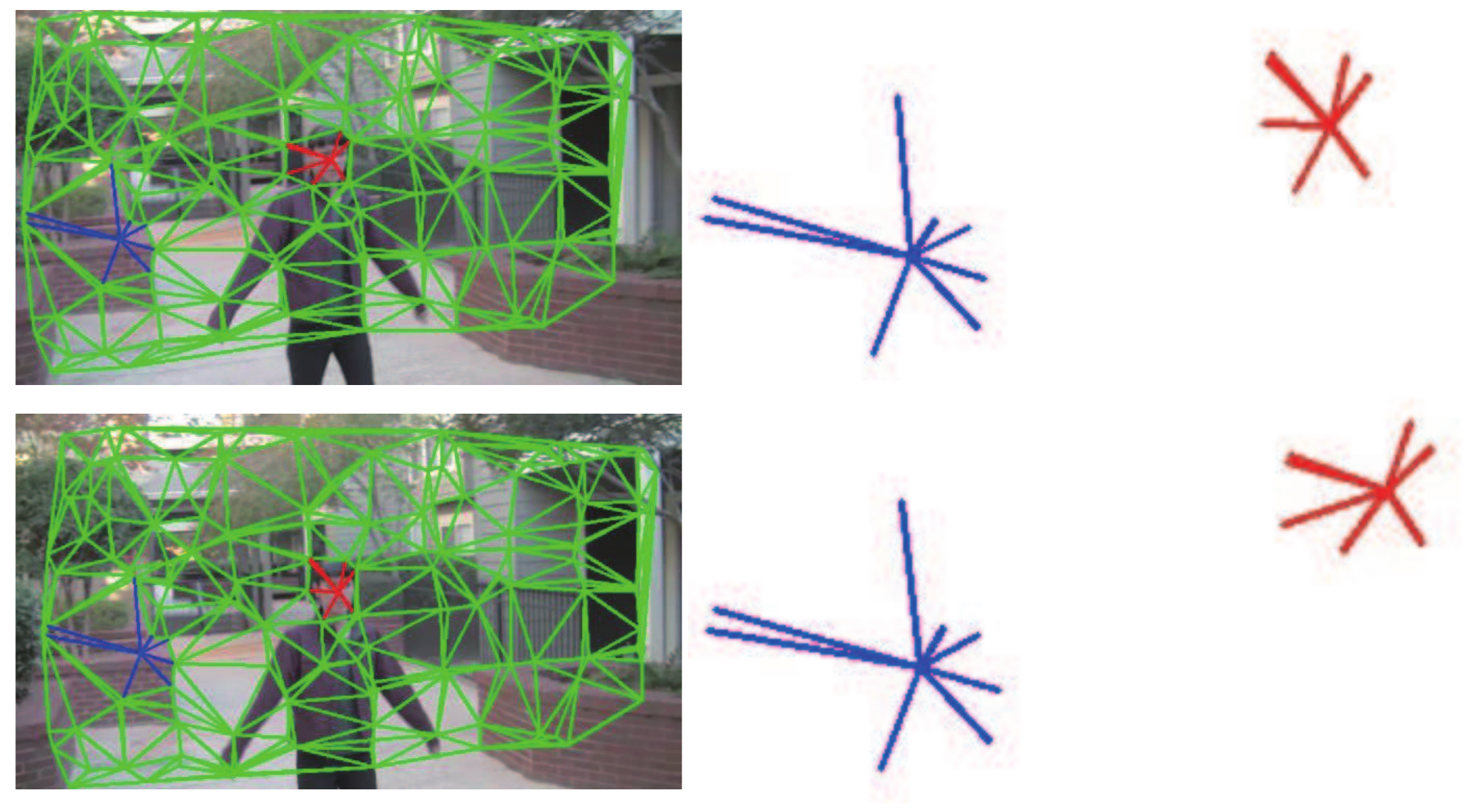}
\caption{Shapes of two point sets at adjacent frames. The blue lines represent the point set belonging to the background while the red lines represent the point set which contains some points belonging to the foreground.}
\label{fig:shaky_inter}
\end{figure}

Given a feature point $i$ in $M_t$ at frame $t$, $i$ and  its $k$ neighboring feature points comprise a set $C_{t,i}=\{C_{t,i,1},...,C_{t,i,k+1}\}$, where $C_{t,i,1}$ is point $i$. As shown in Fig. \ref{fig:shaky_inter}, if all points in $C_{t,i}$ belong to the background, their arrangement at adjacent frames is similar. However, when some points in $C_{t,i}$ belong to the foreground, their arrangement may change much at adjacent frames.  So we take the arrangement change to judge the existence of foreground objects and adjust the weight of $\mathcal{O}_{InterSim}$ in the proposed optimization problem. Specifically, for feature trajectory $i$, we calculate the transformation $H_{t,i}$ between $C_{t,i}$ and $C_{t-1,i}$ by solving the overdetermined equation in (\ref{eq:overdetermined}) with

\newcounter{mytempeqncnt}
\begin{figure*}%???¦Ë?e?????????
\normalsize
\setcounter{mytempeqncnt}{\value{equation}}
\setcounter{equation}{25} %????? څ?????????????????1???????2
\begin{equation}\label{eq:overdetermined}
\underbrace{
\left[\begin{matrix}
 C_{t,i,1}^x &  C_{t,i,1}^y & 1 & 0 & 0 & 0 & -C_{t-1,i,1}^x C_{t,i,1}^x & -C_{t-1,i,1}^x C_{t,i,1}^y     \\
 0 & 0 & 0 &C_{t,i,1}^x &  C_{t,i,1}^y & 1 & -C_{t-1,i,1}^y C_{t,i,1}^x & -C_{t-1,i,1}^y C_{t,i,1}^y      \\
 C_{t,i,2}^x &  C_{t,i,2}^y & 1 & 0 & 0 & 0 & -C_{t-1,i,2}^x C_{t,i,2}^x & -C_{t-1,i,2}^x C_{t,i,2}^y     \\
 0 & 0 & 0 &C_{t,i,2}^x &  C_{t,i,2}^y & 1 & -C_{t-1,i,2}^y C_{t,i,2}^x & -C_{t-1,i,2}^y C_{t,i,2}^y      \\
 \vdots & \vdots & \vdots & \vdots& \vdots& \vdots& \vdots& \vdots  \\
C_{t,i,k+1}^x &  C_{t,i,k+1}^y & 1 & 0 & 0 & 0 & -C_{t-1,i,k+1}^x C_{t,i,k+1}^x & -C_{t-1,i,k+1}^x C_{t,i,k+1}^y     \\
 0 & 0 & 0 &C_{t,i,k+1}^x &  C_{t,i,k+1}^y & 1 & -C_{t-1,i,k+1}^y C_{t,i,k+1}^x & -C_{t-1,i,k+1}^y C_{t,i,k+1}^y      \\
\end{matrix}
\right]}_{\text{\xiaosihao{$A_{t,i}$}}}
\underbrace{
\left[\begin{matrix}
 h_{11}\\
 h_{12}\\
 h_{13}\\
 h_{21}\\
 h_{22}\\
 h_{23}\\
 h_{31}\\
h_{32}\\
\end{matrix}
\right]}_{\text{\xiaosihao{$\beta_{t,i}$}}}=
\underbrace{
\left[\begin{matrix}
C_{t-1,i,1}^x\\
C_{t-1,i,1}^y\\
C_{t-1,i,2}^x\\
C_{t-1,i,2}^y\\
\vdots\\
C_{t-1,i,k+1}^x\\
C_{t-1,i,k+1}^y\\
\end{matrix}
\right]}_{\text{\xiaosihao{$B_{t,i}$}}}.
\end{equation}
\setcounter{equation}{\value{mytempeqncnt}}
\vspace*{4pt} %???????????????
\end{figure*}
%?????????
\setcounter{equation}{26}%????????????????????????????????

\begin{equation}\label{eq:H_3_3}
H_{t,i}=
\left[
\begin{matrix}
 h_{11} & h_{12} & h_{13} \\
 h_{21} & h_{22} & h_{23}  \\
 h_{31} & h_{32} & 1 \\
\end{matrix}
\right].
\end{equation}

\begin{figure*}
\centering\includegraphics[width=0.9\hsize]{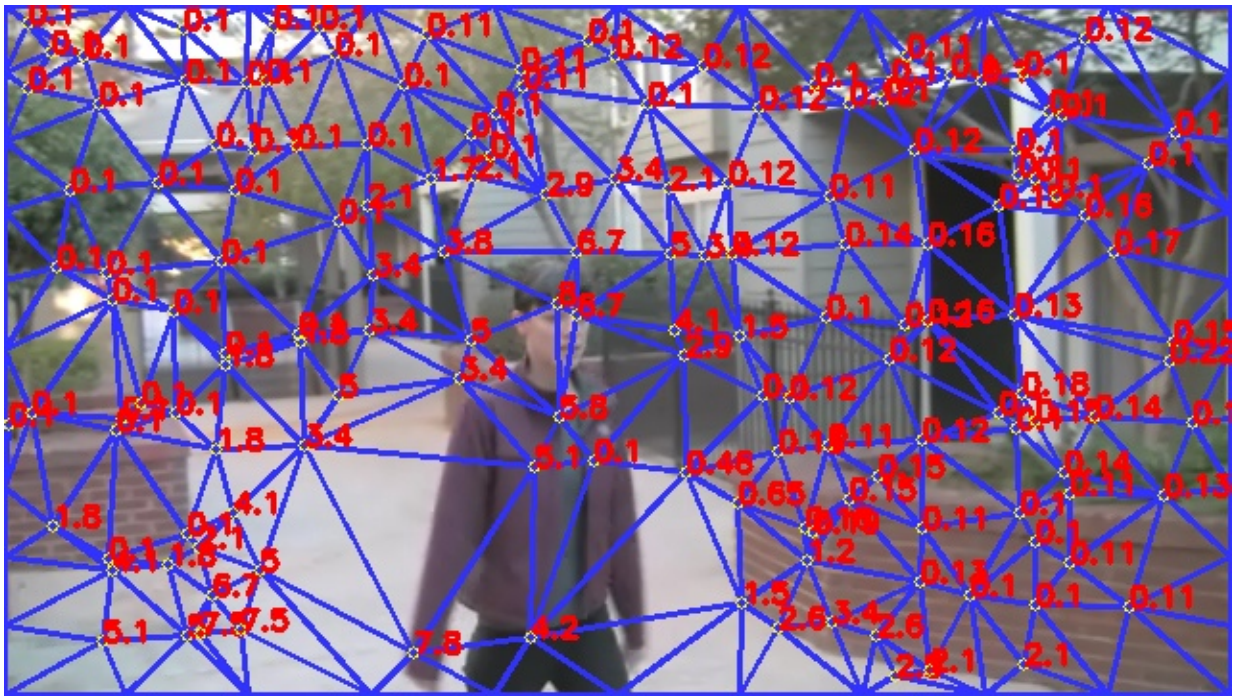}
\caption{ Residual error $LSM_{t,i}$ in an example frame.}
\label{fig:LSM}
\end{figure*}

(\ref{eq:overdetermined}) can be solved by the least square method. (\ref{eq:overdetermined}) can be abbreviated into $A_{t,i}\beta_{t,i}=B_{t,i}$ and provide the solution of $\beta_{t,i}=(A_{t,i}^TA_{t,i})^{-1}A_{t,i}^TB_{t,i}$ with the following residual error
% of $\|A_{t,i}\beta-B_{t,i}\|^2$, which is defined as
\begin{eqnarray}\label{eq:L_grid}
LSM_{t,i}&=&\theta_{t,i}*\|A_{t,i}\beta_{t,i}-B_{t,i}\|^2_2  \\
&=&\theta_{t,i}*\|A_{t,i}(A_{t,i}^TA_{t,i})^{-1}A_{t,i}^TB_{t,i}-B_{t,i}\|^2_2, \nonumber
\end{eqnarray}
where $\theta_{t,i}$ measures the weight of the point set $C_{t,i}$ and is defined as
\begin{equation}
\theta_{t,i}=\frac{1}{k*\rho}\sum_{j=1}^k\|C_{t,i,1}-C_{t,i,j+1}\|,
\end{equation}
where $\rho=(W/\tau+H/\tau)/2$, $W$ and $H$ are the width and height of the video frame, $\tau$ is used to control the block size in the normalization and is set to 10 here.
% ($\tau$ happens to be equal to the number of control points in each edge).
To control the influence of $LSM_{t,i}$ in the proposed optimization problem, $LSM_{t,i}$ is limited into the range of $[0.1,10]$.
Fig. \ref{fig:LSM} shows  $LSM_{t,i}$ in an example frame. It can be seen that large $LSM_{t,i}$ usually appears at the feature points related to  foreground objects while $LSM_{t,i}$ in the background is usually small.

Finally we update $\mathcal{O}_{InterSim}$ into

\begin{equation}
\mathcal{O}_{InterSim} = \sum_{t=1}^T\sum_{p=1}^{K_t}\sum_{j \in \mathcal{V}(p)}\gamma \cdot LSM_{t,j} \|\widehat{Q}_{t,p}^j-\widetilde{Q}_{t,p}^j\|^2, \\
\end{equation}
where $\mathcal{V}(p)$ stands for the set of 3 vertices of triangle $p$.

\section{Experimental Results}
\label{sec:experiments}
\subsection{Dataset and Evaluation Metrics}
\begin{figure*}
\centering\includegraphics[width=\hsize]{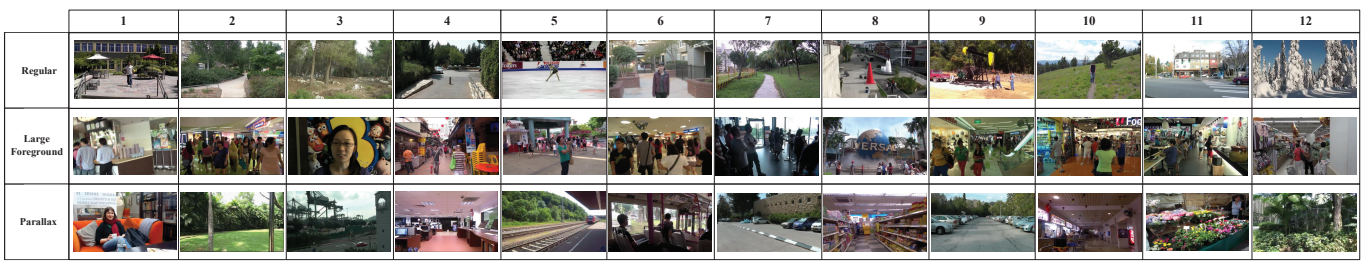}
\caption{Datasets for the experiments. The videos are classified into three categories, including ``Regular'', ``Large foreground'' and ``Parallax''.}
\label{fig:example_images}
\end{figure*}

To verify the performance of our method, we collected 36 typical videos from public datasets\footnote{ \url{http://liushuaicheng.org/SIGGRAPH2013/database.html},\\ \url{http://web.cecs.pdx.edu/~fliu/project/subspace_stabilization/}. }. These videos can be classified into three categories, including {`` Regular''}, { ``Large foreground''} and {``Parallax''}, which are shown in Fig. \ref{fig:example_images}. { ``Regular''} category contains several regular videos recording some common scenes.  { ``Large foreground''} category contains several videos where large foreground objects occupy a large part of video frames and cause serious  discontinuous depth variation. { ``Parallax''} means there is strong parallax in the captured scenes of the videos. The last two categories are more challenging and their experiments confirm that our method can achieve much better performance than previous works.

For quantitative comparison, we consider two evaluation metrics: Stability score and SSIM.

\begin{figure}
\centering\includegraphics[width=0.6\hsize]{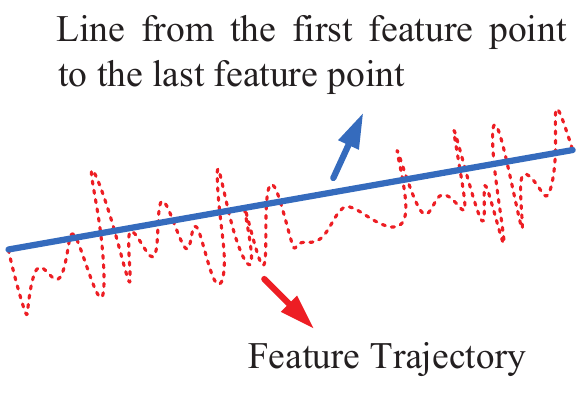}
\caption{The red dotted curve represents a feature trajectory while the
blue line directly connect the start point to the end point of
the feature trajectory. Stability score is defined as the ratio
of length of the blue line to the length of the red dotted curve.}
\label{fig:stability_score}
\end{figure}

{\bf Stability score: }
This metric is taken from \cite{nie2017dynamic}. As shown in Fig. \ref{fig:stability_score}, the red dotted curve represents a feature trajectory, and the blue line  directly connects the start point and the end point of the feature trajectory. Then the stability score is defined as the length of the blue line divided by that of the red dotted curve. The range of stability score is (0,1], and a larger value means a more stable result. Following \cite{ma2019effective}, feature trajectories are extracted by the KLT tracker and are divided into segments of the length of 40 frames. We ignore segments whose lengths are less than 40 frames. The final stability score of a stabilized video is defined as the average stability score of all segments.

{\bf SSIM: }
SSIM (structural similarity) \cite{wang2004image} is a widely-used index to measure the video stabilization performance. This index combines luminance, contrast and structure comparison to calculate the  similarity of two given images. We calculate the SSIM of every two adjacent frames in a video and average them to produce the final SSIM value of a video.

\subsection{Ablation Study}

\begin{figure*}
\centering\includegraphics[width=\hsize]{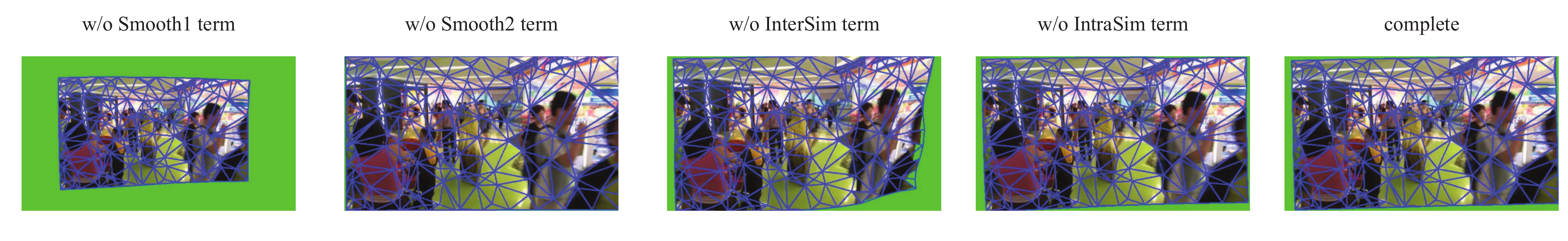}
\caption{Visualization of the stabilized results of the ablation study.
The green background indicates the size of the original frame in the example video.}
%The green regions stand for what are croppedbackground indicates the size of the original frame in the example video.}
\label{fig:ablation}
\end{figure*}

Here we first perform ablation study to evaluate the effectiveness of terms of our proposed optimization framework in (\ref{def:optim}). To analyze the importance of each term in (\ref{eq:optimization}), we remove one term of (\ref{eq:optimization}) at a time and then evaluate the stabilization performance. The quantitative and qualitative results are shown in Fig. \ref{fig:ablation} and Table \ref{tab:ablation}.

Among all terms in (\ref{eq:optimization}), we do not check the influence of the regularization term $\mathcal{O}_{Reg}$ as it is a basic term to make sure that the stabilized trajectories stay close to their original ones and are free of drift. For the other terms, we remove one term from the performance index by setting its weight to zero, i.e., one of  $\alpha$, $\beta$, $\gamma$ and $\varepsilon$ is set to $0$. Table \ref{tab:ablation} shows the quantitative results of the ablation study where both the stability score and SSIM are calculated. Fig. \ref{fig:ablation} shows the visualization results of the ablation study. In Table \ref{tab:ablation}, we find that without $\mathcal{O}_{Smooth1}$, the stability performance is seriously degraded, which confirms that the first-order derivative term is the main one to guarantee the low-frequency characteristics of the stabilized videos. The stability score and SSIM also degrade when the second-order derivative term is removed from the optimization function. According to the analysis in Section \ref{sec:temporal}, removing the second-order derivative term means that feature trajectories are smoothed only with the first-order derivative term and the camera motion is enforced to be static. Thus fast camera movement may yield artifacts and even collapse, which is demonstrated in Fig. \ref{fig:ablation}. $\mathcal{O}_{InterSim}$ also significantly influences the performance. As shown in Fig. \ref{fig:ablation}, due to the lack of the local shape preserving constraint, serious distortion may happen when parallax or large foreground objects exist, which also cause the  non-smoothing of stabilized videos. Table \ref{tab:ablation} also shows that when $\mathcal{O}_{IntraSim}$ is removed, the stabilization performance becomes worse, which illustrates the importance of this term in keeping the similar transformation in the adjacent regions. $\mathcal{O}_{IntraSim}$ can also avoid the trajectory mis-matching and improve the stabilization results.

\begin{table}[!hbp]
  \centering
  \caption{Ablation study of optimization terms. Each row shows the stabilization statistics of three categories,
Regular, Large foreground and parallax, in terms of two metrics, Stability score and SSIM.}
   %\begin{tabularx}{5.8cm}{p{1cm}<{\centering}|p{0.8cm}<{\centering}|p{0.8cm}<{\centering}|p{0.8cm}<{\centering}|p{0.8cm}<{\centering}|p{0.8cm}<{\centering}|p{0.8cm}<{\centering}|}% ????8cm ????????p{}????????????
    \begin{tabular}{|C{1.3cm}|C{0.8cm}|C{0.7cm}|C{0.8cm}|C{0.8cm}|C{0.7cm}|C{0.8cm}|}
    \toprule
    Metrics & \multicolumn{3}{c|}{Stability Score} & \multicolumn{3}{c|}{SSIM} \\
    \midrule
    Optimization Terms & Regular & Large foreground & Parallax & Regular & Large foreground & Parallax \\
    \midrule
    w/o Smooth0 & 0.7998 & 0.7657 & 0.7821 & 0.7045 & 0.7663 & 0.7433 \\
    \midrule
    w/o Smooth1 & 0.8241 & 0.7895 & 0.8268 & 0.7520 & 0.7997 & 0.7965 \\
    \midrule
    w/o InterSim & 0.8407 & 0.8027 & 0.8304 & 0.7609 & 0.8002 & 0.7920 \\
    \midrule
    w/o IntraSim & 0.8514 & 0.8178 & 0.8356 & 0.7664 & 0.8046 & 0.7969 \\
    \midrule
    complete & 0.8589 & 0.8287 & 0.8541 & 0.7750 & 0.8196 & 0.8033 \\
    \bottomrule
    \end{tabular}%
  \label{tab:ablation}%
\end{table}%

\subsection{Comparison with Previous Methods}

We compare our method with several state-of-the-art methods, including Subspace \cite{liu2011subspace}, Meshflow \cite{liu2016meshflow}, Deep learning \cite{wang2018deep}, Effective \cite{ma2019effective} and Global \cite{zhang2015global}. Subspace is implemented in Adobe Premiere stabilizer. The codes of  Meshflow \cite{liu2016meshflow}, Deep learning \cite{wang2018deep} and Effective \cite{ma2019effective} were found at https://github.com/sudheerachary/Mesh-Flow-Video-Stabilization, https://github.com/cxjyxxme/deep-online-video-stabilization-deploy and https://github.com/705062791/TVCG-Video-Stabilization-via-joint-Trajectory-Smoothing-and-frame-warping. Thanks to the authors of\cite{zhang2015global}, who provided us with the binary implementation of Global.

\begin{figure*}
\centering\includegraphics[width=\hsize]{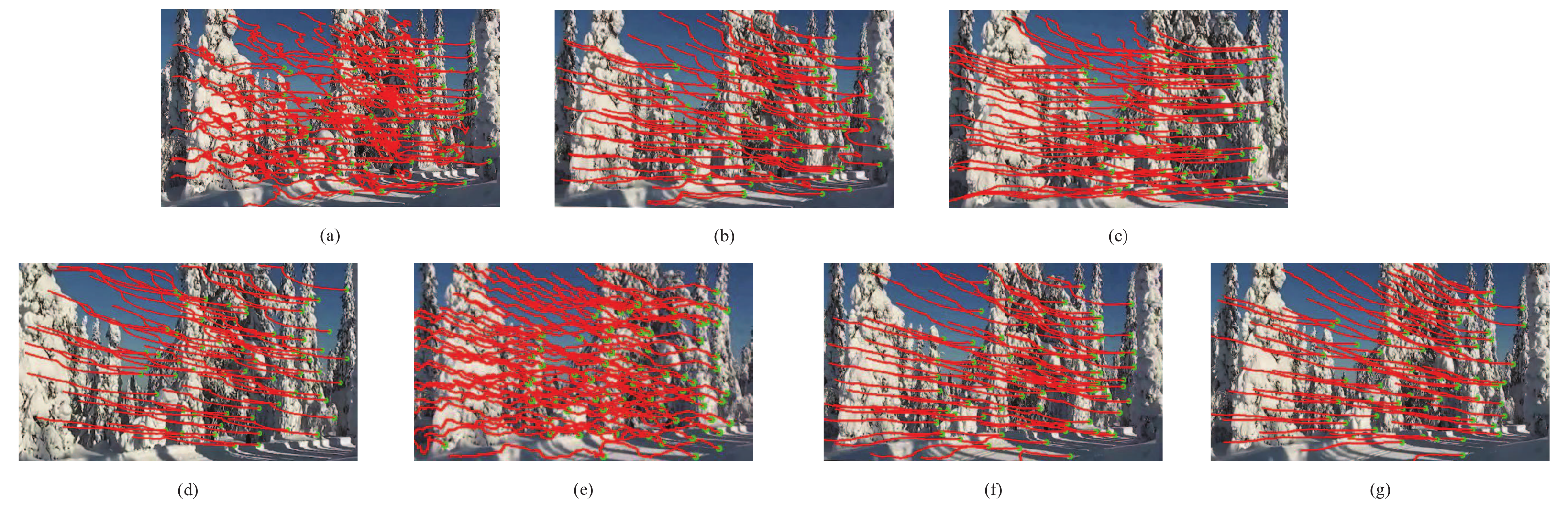}
\caption{Comparison of stabilized feature trajectories. (a) Input video. (b) Effective \cite{ma2019effective}. (c) Meshflow \cite{liu2016meshflow}. (d) Subspace \cite{liu2011subspace}. (e) Deep learning \cite{wang2018deep}. (f) Global \cite{zhang2015global}. (g) Proposed.}
\label{fig:trj_compare}
\end{figure*}

\begin{figure*}
\centering\includegraphics[width=\hsize]{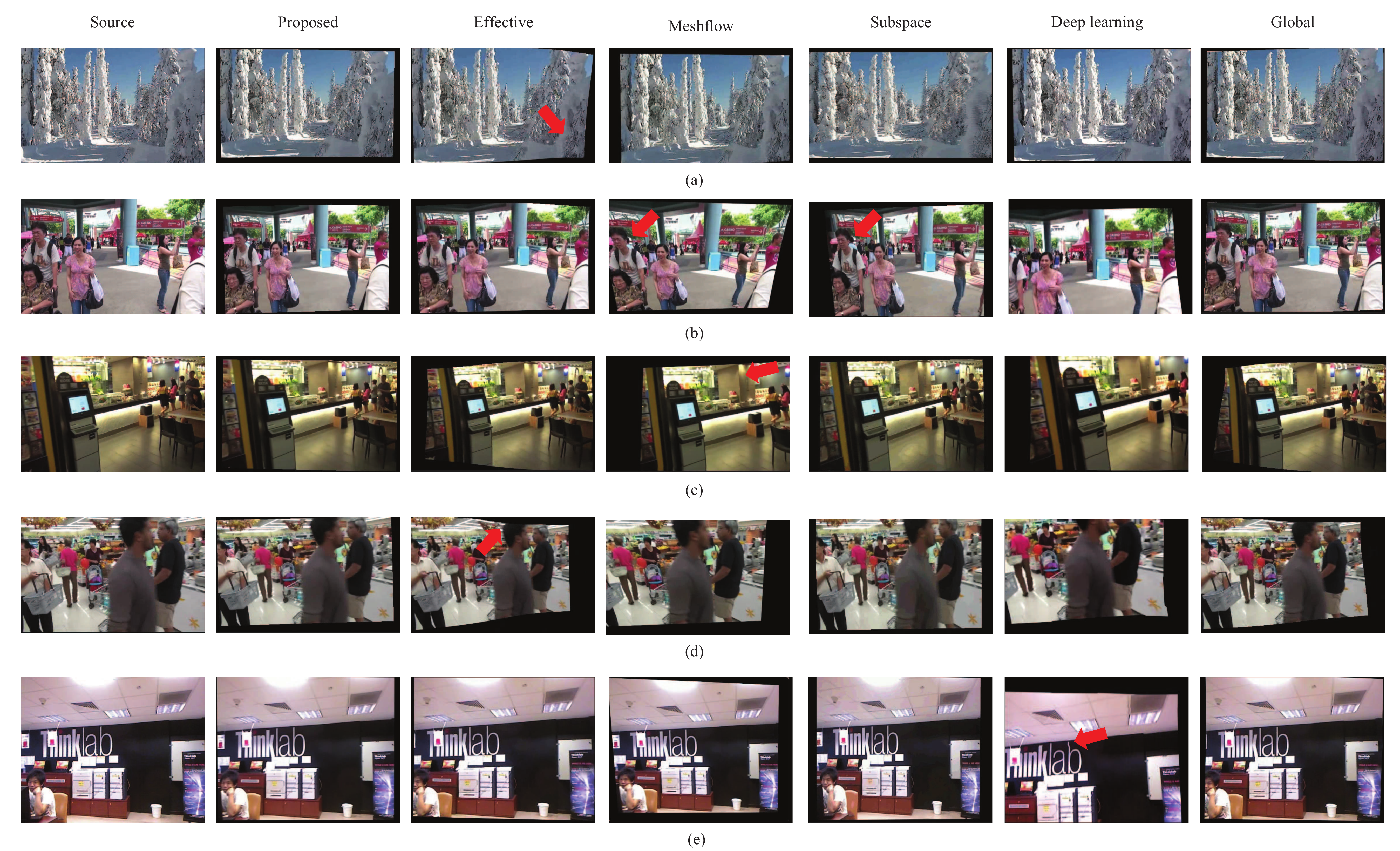}
\caption{Comparison of stabilization results by the proposed method, Effective \cite{ma2019effective}, Meshflow \cite{liu2016meshflow}, Subspace \cite{liu2011subspace}, Deep learning \cite{wang2018deep} and Global \cite{zhang2015global}. (a)-(e) show 5 examples of sampled frames from shaky videos with different motions.}
\label{fig:distortion}
\end{figure*}

Here we first compare our method with these methods through some qualitative results, which are shown in Fig. \ref{fig:trj_compare} and Fig. \ref{fig:distortion}. Fig. \ref{fig:trj_compare} shows feature trajectories of an example shaky video and the corresponding stabilized videos. The Deep learning method stabilizes videos without using future frames. Therefore, compared with other methods, its stabilization effect is the worst. Although the Effective method reduces most of high-frequency shakiness, we can still catch many visible low-frequency footstep motions. Therefore, its trajectories are not good enough. The results of the Subspace, Meshflow and Global methods eliminate almost all high-frequency shakiness, and generate similar stabilization effects. The results of the proposed method show significant improvement, which is confirmed by the smoothness of the stabilized trajectories.
%Therefore, according to the shape of feature trajectories, the proposed method can provide the smoothest stabilized motions.

In addition to stability, noticeable visual distortion should also be avoided in stabilized videos. Fig. \ref{fig:distortion} shows some example frames in several stabilized videos by different methods. The Subspace method performs motion compensation through smoothing long feature trajectories with low-rank constraints. As shown by Fig. \ref{fig:distortion}(b), when there are not enough long feature trajectories, for example, in the case of fast moving foreground objects, Subspace may fail and result in serious distortion. The Meshflow and Global methods discard foreground feature trajectories and estimate the camera motion with only the background feature trajectories. However, separating foreground trajectories from background ones is a difficult task, especially in large foreground scenes or strong parallax. So these methods may suffer performance degradation in challenging videos, as shown by  Fig. \ref{fig:distortion}(b) and \ref{fig:distortion}(c). The Deep learning method stabilizes videos without future frames and regresses one homography for each of fixed grids. The inaccurate estimation and weak constraint for local shape preserving may incur noticeable distortion, which is illustrated in Fig. \ref{fig:distortion}(e). The Effective method estimates the camera motion based on all feature trajectories. However, this method divides frames into fixed mesh grids of the same size and extract a similar number of feature trajectories in each mesh grid, which is not reasonable for complicated scenes, such as large foreground and parallax.  Fig. \ref{fig:distortion}(a) and \ref{fig:distortion}(d) show that this method may cause distortion in regions with  rich contents or large depth variation.

The quantitative comparison on all test videos is also performed. Table \ref{tab:stability} and Table \ref{tab:ssim} show Stability score and SSIM generated by different methods on  all test videos. In terms of Stability score, the proposed method outperforms all other methods on most videos in the ``Regular'' category. Only in two videos, the Subspace method gets the highest Stability score, which is slightly better than that of the proposed method. However, the Subspace method gets worse results than some other methods in other videos.
More significant performance improvement of the proposed method can be observed in the other two complicated categories, ``Large foreground'' and ``Parallax''. Our method gets much higher Stability scores on most videos. Note that the compared methods, except the Subspace method, take fixed meshes in stabilizing videos.  For the scenes containing large foreground objects or parallax, these methods may suffer serious performance degradation while our method can get much better results with the adaptively adjusted mesh.
 Moreover, the proposed method takes both background feature trajectories and foreground feature trajectories to smooth the camera motion and may be hardly disturbed by the active motion of the foreground objects. Thus the proposed method achieves higher Stability scores in most videos than the methods which discard foreground feature trajectories. So our method is the most robust among all methods.
Similar SSIM comparison results can also be observed in Table \ref{tab:ssim}. The proposed method gets much better stability score and SSIM in most videos than other methods. Since stability score measures the stability of local feature trajectories and SSIM measures the global similarity of adjacent frames in the stabilized videos, the results confirm the effectiveness of our method.
For more thorough comparison, a supplemental demo is provided at \url{http://home.ustc.edu.cn/~zmd1992/content-aware-stabilization.html}.

\begin{table*}[htbp]
  \centering
  \caption{Stability scores of the proposed method and some state-of-the-art methods on three video categories, including Regular, Large foreground and Parallax. ``/'' means the corresponding method fails to stabilize that video.}
    \begin{tabular}{|c|c|c|c|c|c|r|c|c|c|c|c|r|c|}
    \toprule
    \multirow{7}[14]{*}{Regular} & Methods & 1     & 2     & 3     & 4     & \multicolumn{1}{c|}{5} & 6     & 7     & 8     & 9     & 10    & \multicolumn{1}{c|}{11} & 12 \\
\cmidrule{2-14}          & Effective \cite{ma2019effective} & 0.7529 & 0.7483 & 0.9292 & 0.6516 & \multicolumn{1}{c|}{0.7161} & 0.8201 & 0.6965 & 0.7141 & 0.7413 & 0.624 & \multicolumn{1}{c|}{0.8022} & 0.9036 \\
\cmidrule{2-14}          & Meshflow \cite{liu2016meshflow} & 0.7295 & 0.8298 & 0.9260 & 0.7272 & \multicolumn{1}{c|}{0.8137} & 0.8247 & 0.6564 & 0.7069 & 0.7813 & 0.5147 & \multicolumn{1}{c|}{0.8061} & 0.9151 \\
\cmidrule{2-14}          & Subspace \cite{liu2011subspace} & 0.6633 & \textbf{0.9049} & 0.9515 & 0.8328 & \multicolumn{1}{c|}{0.6552} & 0.4191 & 0.7281 & 0.5866 & \textbf{0.8526} & 0.4527 & \multicolumn{1}{c|}{0.6165} & 0.8270 \\
\cmidrule{2-14}          & Deep learning \cite{wang2018deep}    & 0.4056 & 0.5601 & 0.6825 & 0.4454 & \multicolumn{1}{c|}{0.6679} & 0.6007 & 0.4299 & 0.2597 & 0.5572 & 0.4200  & \multicolumn{1}{c|}{0.5353} & 0.7781 \\
\cmidrule{2-14}          & Global \cite{zhang2015global}  & 0.8282 & 0.8791 & 0.9650 & 0.7779 & \multicolumn{1}{c|}{0.7347} & 0.8441 & 0.6432 & 0.7739 & 0.7915 & 0.6808 & \multicolumn{1}{c|}{0.7846} & 0.928 \\
\cmidrule{2-14}          & Proposed & \textbf{0.8559} & 0.8974 & \textbf{0.9686} & \textbf{0.8372} & \multicolumn{1}{c|}{\textbf{0.8578}} & \textbf{0.8513} & \textbf{0.8096} & \textbf{0.8138} & 0.8334 & \textbf{0.7810} & \multicolumn{1}{c|}{\textbf{0.8578}} & \textbf{0.9437} \\
    \midrule
    \multirow{7}[14]{*}{\makecell[c]{Large \\ foreground}} & Methods & 1     & 2     & 3     & 4     & \multicolumn{1}{c|}{5} & 6     & 7     & 8     & 9     & 10    & \multicolumn{1}{c|}{11} & 12 \\
\cmidrule{2-14}          & Effective \cite{ma2019effective}  & 0.8189 & 0.7241 & \multicolumn{1}{r|}{0.5980} & 0.7169 & \multicolumn{1}{c|}{0.8049} & 0.7149 & 0.7222 & 0.8387 & 0.7253 & 0.7322 & \multicolumn{1}{c|}{0.8257} & 0.7782 \\
\cmidrule{2-14}          & Meshflow \cite{liu2016meshflow} & 0.8495 & 0.7197 & \multicolumn{1}{r|}{0.6082} & 0.7326 & \multicolumn{1}{c|}{0.7542} & 0.7404 & 0.8167 & 0.8376 & 0.7442 & 0.7427 & \multicolumn{1}{c|}{0.8198} & 0.7891 \\
\cmidrule{2-14}          & Subspace \cite{liu2011subspace} & 0.5884 & 0.6768 & \multicolumn{1}{r|}{0.4784} & 0.6685 & \multicolumn{1}{c|}{0.6525} & 0.6130 & 0.7071 & 0.5659 & 0.3239 & 0.6130 & \multicolumn{1}{c|}{0.7747} & 0.6698 \\
\cmidrule{2-14}          & Deep learning \cite{wang2018deep}    & 0.3437 & 0.3832 & \multicolumn{1}{r|}{0.4764} & 0.4690 & \multicolumn{1}{c|}{0.4371} & 0.4760 & 0.6218 & 0.5000   & 0.5242 & 0.6430 & \multicolumn{1}{c|}{0.5544} & 0.4960 \\
\cmidrule{2-14}          & Global \cite{zhang2015global}  & 0.7575 & 0.4076 & \multicolumn{1}{r|}{0.6167} & 0.6213 & \multicolumn{1}{c|}{0.7679} & 0.6505 & \textbf{0.8395} & 0.8633 & 0.6253 & 0.6953 & \multicolumn{1}{c|}{0.7827} & 0.7411 \\
\cmidrule{2-14}          & PROPOSED & \textbf{0.8901} & \textbf{0.7903} & \multicolumn{1}{r|}{\textbf{0.6305}} & \textbf{0.8249} & \multicolumn{1}{c|}{\textbf{0.8652}} & \textbf{0.8064} & 0.8369 & \textbf{0.8893} & \textbf{0.8161} & \textbf{0.8422} & \multicolumn{1}{c|}{\textbf{0.8841}} & \textbf{0.8690} \\
    \midrule
    \multirow{7}[14]{*}{Parallax} & Methods & 1     & 2     & {3} & 4     & \multicolumn{1}{c|}{5} & 6     & 7     & 8     & 9     & 10    & \multicolumn{1}{c|}{11} & 12 \\
\cmidrule{2-14}          & Effective \cite{ma2019effective}  & 0.8315 & 0.8498 & 0.7044 & 0.7447 & 0.8544 & 0.7371 & 0.6895 & \textbf{0.9412} & 0.7450 & 0.7600  & 0.8716 & 0.4855 \\
\cmidrule{2-14}          & Meshflow \cite{liu2016meshflow} & 0.8082 & 0.9039 & 0.7170 & 0.7162  & \multicolumn{1}{l|}{/} & /     & 0.6729 & 0.6398 & 0.8021 & 0.7957 & 0.8682 & 0.5898 \\
\cmidrule{2-14}          & Subspace \cite{liu2011subspace} & 0.8112 & \textbf{0.9125} & 0.4767 & 0.6697 & 0.8854 & 0.8035 & 0.8272 & 0.9174 & 0.8963 & 0.7942 & \textbf{0.9170} & 0.5100 \\
\cmidrule{2-14}          & Deep learning \cite{wang2018deep}    & 0.6140 & 0.7561 & 0.5496 & 0.6442 & 0.5768 & 0.5684 & 0.5246 & 0.5950 & 0.4338 & 0.4314 & 0.7110 & 0.3455 \\
\cmidrule{2-14}          & Global \cite{zhang2015global}  & 0.8089 & 0.5740 & 0.6862 & 0.8052 & 0.7635 & 0.7453 & 0.7519 & /     & 0.8675 & 0.8204 & 0.8905 & 0.6501 \\
\cmidrule{2-14}          & PROPOSED & \textbf{0.8648} & 0.9111 & \textbf{0.7663} & \textbf{0.8549} & \textbf{0.8983} & \textbf{0.8216} & \textbf{0.8291} & 0.9238 & \textbf{0.9076} & \textbf{0.8444} & 0.9059 & \textbf{0.7208} \\
    \bottomrule
    \end{tabular}%
  \label{tab:stability}%
\end{table*}%

\begin{table*}[htbp]
  \centering
  \caption{SSIMs of the proposed method and some state-of-the-art methods on three video categories, including Regular, Large foreground and Parallax. ``/'' means the corresponding method fails to stabilize that video.}
    \begin{tabular}{|c|c|c|c|c|c|c|c|c|c|c|c|c|c|}
    \toprule
    \multirow{7}[14]{*}{Regular} & Methods & 1     & 2     & 3     & 4     & 5     & 6     & 7     & 8     & 9     & 10    & 11    & 12 \\
\cmidrule{2-14}          & Effective \cite{ma2019effective} & 0.8026 & 0.6804 & 0.5647 & 0.8195 & 0.8004 & 0.6913 & 0.7781 & 0.8874 & 0.8755 & 0.7693 & 0.7507 & 0.5814 \\
\cmidrule{2-14}          & Meshflow \cite{liu2016meshflow} & 0.8260 & 0.6815 & 0.4919 & 0.8423 & \textbf{0.8363} & 0.6986 & 0.8084 & 0.9042 & 0.8690 & 0.7440 & 0.7658 & 0.5257 \\
\cmidrule{2-14}          & Subspace \cite{liu2011subspace} & 0.5637 & \textbf{0.6895} & 0.4587 & 0.8443 & 0.8328 & 0.6159 & 0.7608 & 0.6056 & 0.8604 & 0.6546 & 0.5462 & 0.4679 \\
\cmidrule{2-14}          & Deep learning \cite{wang2018deep}    & 0.7133 & 0.6638 & 0.5597 & 0.7666 & 0.7371 & 0.6477 & 0.6627 & 0.7813 & 0.8636 & 0.7037 & 0.6656 & 0.5831 \\
\cmidrule{2-14}          & Global \cite{zhang2015global}  & 0.7943 & 0.6101 & 0.4388 & 0.7752 & 0.7912 & 0.6569 & 0.7403 & 0.8764 & 0.8558 & 0.7192 & 0.7518 & 0.5566 \\
\cmidrule{2-14}          & PROPOSED & \textbf{0.8447} & 0.6712 & \textbf{0.6171} & \textbf{0.8461} & 0.8009 & \textbf{0.7434} & \textbf{0.8141} & \textbf{0.9240} & \textbf{0.9083} & \textbf{0.8004} & \textbf{0.8301} & \textbf{0.6000} \\
    \midrule
    \multirow{7}[14]{*}{\makecell[c]{Large \\ foreground}} & Methods & 1     & 2     & 3     & 4     & 5     & 6     & 7     & 8     & 9     & 10    & 11    & 12 \\
\cmidrule{2-14}          & Effective \cite{ma2019effective} & 0.6771 & 0.7365 & 0.8261 & 0.6815 & 0.7213 & 0.7657 & 0.7057 & 0.6525 & 0.7579 & 0.7096 & 0.7135 & 0.6774 \\
\cmidrule{2-14}          & Meshflow \cite{liu2016meshflow} & 0.7450 & 0.7978 & 0.8545 & 0.7423 & 0.7596 & 0.8132 & 0.7470 & 0.6852 & 0.8275 & 0.7746 & 0.782 & 0.7458 \\
\cmidrule{2-14}          & Subspace \cite{liu2011subspace} & 0.4738 & 0.6359 & 0.6890 & 0.5127 & 0.623 & 0.7176 & 0.6144 & 0.5072 & 0.4916 & 0.5212 & 0.5182 & 0.5679 \\
\cmidrule{2-14}          & Deep learning \cite{wang2018deep}    & 0.5223 & 0.5935 & 0.7977 & 0.5830 & 0.6056 & 0.6747 & 0.6396 & 0.5525 & 0.6337 & 0.6409 & 0.6336 & 0.5815 \\
\cmidrule{2-14}          & Global \cite{zhang2015global}  & 0.7270 & 0.5711 & 0.8164 & 0.7101 & 0.7502 & 0.7953 & 0.7191 & 0.6639 & 0.7966 & 0.7593 & 0.7625 & 0.7360 \\
\cmidrule{2-14}          & PROPOSED & \textbf{0.7999} & \textbf{0.8379} & \textbf{0.8573} & \textbf{0.7955} & \textbf{0.8226} & \textbf{0.8638} & \textbf{0.7985} & \textbf{0.7540} & \textbf{0.8588} & \textbf{0.8266} & \textbf{0.8260} & \textbf{0.7948} \\
    \midrule
    \multirow{7}[14]{*}{Parallax} & Methods & 1     & 2     & 3     & 4     & 5     & 6     & 7     & 8     & 9     & 10    & 11    & 12 \\
\cmidrule{2-14}          & Effective \cite{ma2019effective} & 0.7351 & 0.6109 & 0.8721 & 0.7459 & 0.7744 & 0.8065 & 0.8788 & 0.5493 & 0.8822 & 0.9048 & 0.6441 & 0.8677 \\
\cmidrule{2-14}          & Meshflow \cite{liu2016meshflow} & 0.7420 & 0.6297 & 0.9006 & 0.7219     & /     &   /    & 0.8649 & 0.5550 & 0.8903 & \textbf{0.9299} & 0.6312 & 0.8850 \\
\cmidrule{2-14}          & Subspace \cite{liu2011subspace} & 0.6877 & 0.4779 & 0.8544 & 0.7544 & 0.7769 & 0.7889 & 0.8949 & 0.5324 & 0.8987 & 0.9251 & 0.6242 & 0.7845 \\
\cmidrule{2-14}          & Deep learning \cite{wang2018deep}    & 0.7261 & 0.5278 & 0.8431 & 0.6896 & 0.7872 & 0.6933 & 0.7749 & 0.4332 & 0.7882 & 0.8555 & 0.5920 & 0.8531 \\
\cmidrule{2-14}          & Global \cite{zhang2015global}  & 0.7143 & \textbf{0.7005} & 0.8793 & 0.7428 & 0.7879 & 0.7764 & 0.8614 & /     & 0.8640 & 0.8660 & 0.6423 & 0.8243 \\
\cmidrule{2-14}          & PROPOSED & \textbf{0.7740} & 0.6732 & \textbf{0.9141} & \textbf{0.7808} & \textbf{0.8278} & \textbf{0.8441} & \textbf{0.8998} & \textbf{0.5667} & \textbf{0.9021} & 0.9130 & \textbf{0.6558} & \textbf{0.8889} \\
    \bottomrule
    \end{tabular}%
  \label{tab:ssim}%
\end{table*}%

\subsection{Parameter Sensitivity}

Here we investigate the parameter sensitivity of the proposed method. Since the weight of $\mathcal{O}_{Reg}$ is normalized to 1, there are four parameters, including $\alpha$, $\beta$, $\gamma$ and $\varepsilon$, which are the weights of $\mathcal{O}_{Smooth1}$, $\mathcal{O}_{Smooth2}$, $\mathcal{O}_{InterSim}$ and $\mathcal{O}_{IntraSim}$, respectively. For one term, a larger weight implies more importance.

\begin{figure*}
\centering\includegraphics[width=0.75\hsize]{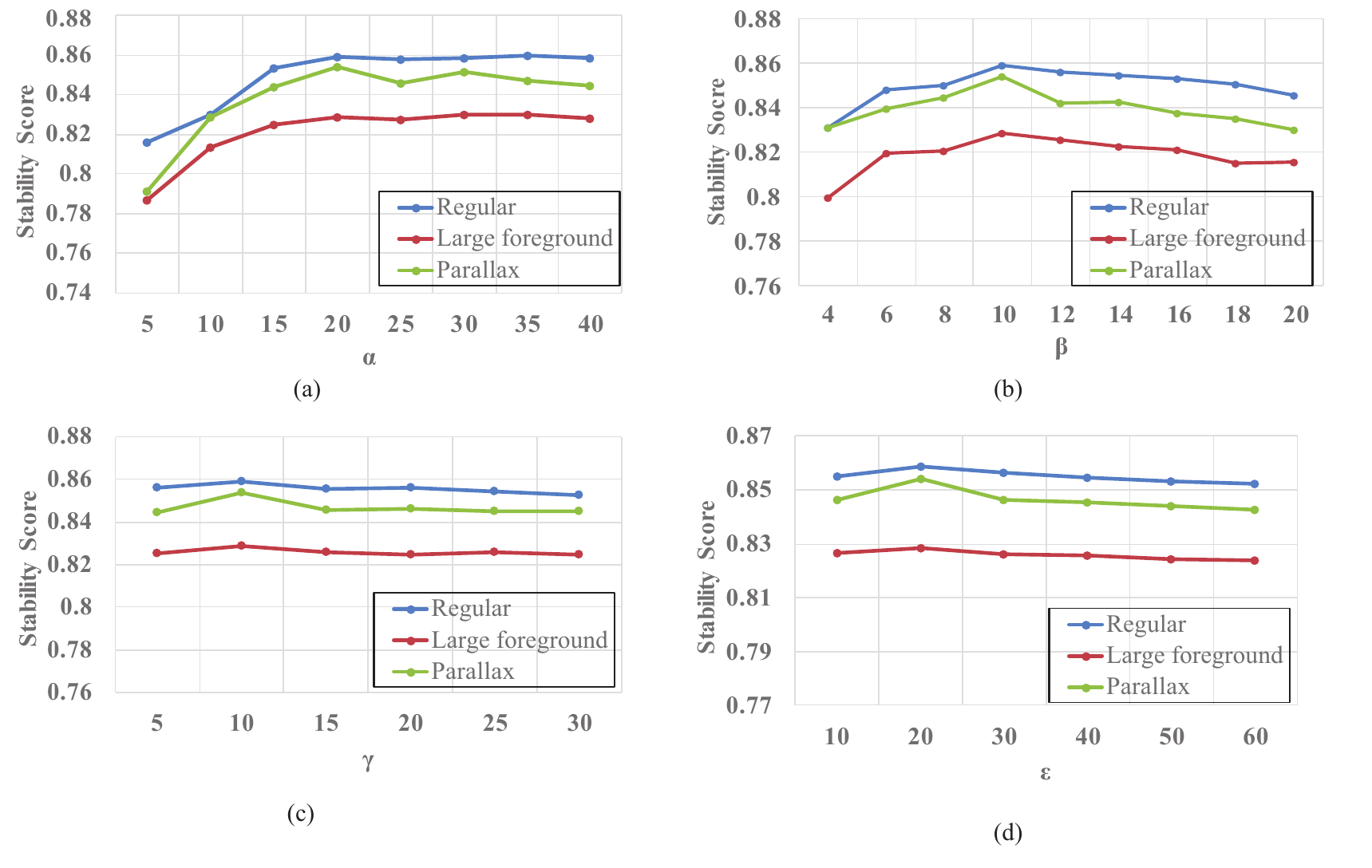}
%\vspace{-0.2in}
\caption{Average stability scores on three categories, i.e., Regular, Large foreground and Parallax, with the variation of four weighting parameters. \,\,\,\,\,\,\,\,\,\,\,\,\,\,\,\,\,\,\,\,\,\,\,\,$\textrm{(a) Various }\alpha$; (b) Various $\beta$; (c) Various $\gamma$; (d) Various $\varepsilon$. }
\label{fig:sensitivity}
%\vspace{-0.2in}
\end{figure*}

Each time we change one weight, fix the others and calculate the average stability scores on three categories to quantitatively measure its sensitivity.
\begin{itemize}
\item Fig. \ref{fig:sensitivity}(a) shows the influence of $\alpha$, where $\alpha$ varies from 5 to 40. We find that when $\alpha$ is smaller than 20, the performance becomes obviously better as $\alpha$ increases. When $\alpha$ is large than 20, the stability score will slightly increase. However, according to the previous analysis, too large $\alpha$ will result in serious collapse risk. So $\alpha$ is set to 20.
\item As shown in Fig. \ref{fig:sensitivity}(b), when $\beta$ increases from 4 to 20, the stability score first increases and then decreases on all three categories. Note that both the increasing amount and the decreasing amount of the stability score are small. So the stabilization performance is not much sensitive to $\beta$. $\beta$ is set to 10.
\item The results under various $\gamma$ and $\varepsilon$ are shown in Fig. \ref{fig:sensitivity}(c) and Fig. \ref{fig:sensitivity}(d). When $\gamma$ and $\varepsilon$ increase, the stability scores first slightly increase and then slowly decrease, which means the stabilization performance is not sensitive to $\gamma$ or $\varepsilon$.  Note that increasing $\gamma$ and $\varepsilon$ is equivalent to decreasing the weights of two smoothing terms, i.e., $\mathcal{O}_{Smoohth1}$ and $\mathcal{O}_{Smoohth2}$. $\gamma$ and $\varepsilon$ are set to 10 and 20, respectively.
\end{itemize}

\section{Conclusion}
\label{sec:conclusion}
In this paper, we propose an adaptively meshed video stabilization method. Different from previous works, which stabilize shaky videos based on fixed meshes, we propose a novel stabilization method by considering the contents of the video and adaptively dividing each frame into a flexible number of mesh grids. The adaptive blocking strategy is realized by the Delaunay triangulation of feature trajectories and all mesh grids are triangles with various sizes. The stabilized views of all triangles in the triangle mesh are calculated through solving a two-stage optimization problem. The proposed method takes both foreground feature trajectories and background feature trajectories for  precise jitter estimation.  To further enhance the robustness of our method, we propose two adaptive weighting mechanisms to improve the spatial and temporal adaptability. Experiments on several public videos confirm the performance superiority of our method.

There are two main limitations of the proposed method. The first one is its computational burden. Our method takes an average of 0.3s to process one frame, which is slower than some previous works, such as Meshflow \cite{liu2016meshflow} and Global \cite{zhang2015global}. The further acceleration of the proposed method will be pursued in the future. The other limitation is also common for other traditional video stabilization methods, i.e., the video stabilization is built upon feature trajectories and may be fragile for low-quality videos with non-texture scenes where it is hard to extract an enough number of reliable feature trajectories. Some deep learning based methods have been proposed to solve this problem, such as \cite{wang2018deep} and \cite{xu2018deep}. We will keep eyes on that direction.

\bibliographystyle{IEEEtran}
\bibliography{refer}

\iffalse
\begin{IEEEbiography}{Michael Shell}
Biography text here.
\end{IEEEbiography}

\begin{IEEEbiographynophoto}{John Doe}
Biography text here.
\end{IEEEbiographynophoto}

%\newpage

\begin{IEEEbiographynophoto}{Jane Doe}
Biography text here.
\end{IEEEbiographynophoto}
\fi

%\vfill


\end{document}